\documentclass[sigconf]{aamas} 

\usepackage{balance} 
\usepackage{amsmath}
\usepackage{subfigure}
\usepackage{longtable}
\usepackage{balance}

\setcopyright{ifaamas}
\acmConference[AAMAS '21]{Proc.\@ of the 20th International Conference on Autonomous Agents and Multiagent Systems (AAMAS 2021)}{May 3--7, 2021}{Online}{U.~Endriss, A.~Now\'{e}, F.~Dignum, A.~Lomuscio (eds.)}
\copyrightyear{2021}
\acmYear{2021}
\acmDOI{}
\acmPrice{}
\acmISBN{}

\acmSubmissionID{86}

\title[AAMAS-2021 Formatting Instructions]{Structured Diversification Emergence via Reinforced Organization Control and Hierarchical Consensus Learning}

\author{Wenhao Li}
\affiliation{
  \institution{East China Normal University}
  \city{Shanghai, China}}
\email{52194501026@stu.ecnu.edu.cn}

\author{Xiangfeng Wang}
\authornote{Corresponding authors}
\affiliation{
  \institution{East China Normal University and SRIAS, Shanghai, China}}
\email{xfwang@cs.ecnu.edu.cn}

\author{Bo Jin}
\authornotemark[1]
\affiliation{
  \institution{East China Normal University and SRIAS, Shanghai, China}}
\email{bjin@cs.ecnu.edu.cn}

\author{Junjie Sheng}
\affiliation{
  \institution{East China Normal University}
  \city{Shanghai, China}}
\email{52194501003@stu.ecnu.edu.cn}

\author{Yun Hua}
\affiliation{
  \institution{East China Normal University}
  \city{Shanghai, China}}
\email{52194501002@stu.ecnu.edu.cn}

\author{Hongyuan Zha}
\affiliation{
  \institution{School of Data Science and AIRS, The Chinese University of Hong Kong, Shenzhen, China. zhahy@cuhk.edu.cn}}

\begin{abstract}
When solving a complex task, humans will spontaneously form teams and to complete different parts of the whole task, respectively.
Meanwhile, the cooperation between teammates will improve efficiency.
However, for current cooperative MARL methods, the cooperation team is constructed through either heuristics or end-to-end blackbox optimization.
In order to improve the efficiency of cooperation and exploration, we propose a structured diversification emergence MARL framework named {\sc{Rochico}} based on reinforced organization control and hierarchical consensus learning.
{\sc{Rochico}} first learns an adaptive grouping policy through the organization control module, which is established by independent multi-agent reinforcement learning.
Further, the hierarchical consensus module based on the hierarchical intentions with consensus constraint is introduced after team formation.
Simultaneously, utilizing the hierarchical consensus module and a self-supervised intrinsic reward enhanced decision module, the proposed cooperative MARL algorithm {\sc{Rochico}} can output the final diversified multi-agent cooperative policy.
All three modules are organically combined to promote the structured diversification emergence.
Comparative experiments on four large-scale cooperation tasks show that {\sc{Rochico}} is significantly better than the current SOTA algorithms in terms of exploration efficiency and cooperation strength.
\end{abstract}


\newcommand{\BibTeX}{\rm B\kern-.05em{\sc i\kern-.025em b}\kern-.08em\TeX}

\begin{document}


\pagestyle{fancy}
\fancyhead{}


\maketitle 


\section{Introduction}
Multi-agent reinforcement learning (MARL) has been widely used and achieve fantastic performance in many application fields, like multiplayer games~\citep{peng2017multiagent,li2020f2a2}, swarm robot control~\citep{matignon2012coordinated} etc.
Most of the current MARL algorithms follow the centralized training and decentralized execution (CTDE, \citep{oliehoek2008optimal}) framework.
In the centralized training phase, a decentralized policy needs to be learned for each agent through sharing local observations, parameters, or gradients among agents.
However, 
these CTDE-based MARL algorithms have to consider each agent as an independent individual during the training procedure.
Although these individuals can transmit information explicitly or implicitly to achieve collaboration, most of them are usually learned through an end-to-end blackbox scheme, which raises the difficulty to obtain meaningful communication protocols~\citep{tian2018learning}.
This makes it difficult for multi-agents to explore and collaborate effectively.

When humans perform tasks in an unknown environment, diverse teams instead of individuals, are usually used as the basic unit to make up for limited individual abilities.
As for the nonstationary of the external environment and the difficulty of the related task, the team can be restructured accordingly.
In terms of MARL, this means that the agents must have the ability to dynamically team-up. 
To improve the efficiency of exploration in an unknown environment, the behavior of different teams need to be sufficiently diverse; 
Further considering the capacity limitations of a single agent, the agents from the same team need to cooperate closely to improve the efficiency of task completion.
In this paper, we call this inter-team diversification and intra-team cooperation ability the \textit{structured diversification emergence}.

\begin{figure*}[htbp]
  \centering
  \includegraphics[width=0.9\textwidth]{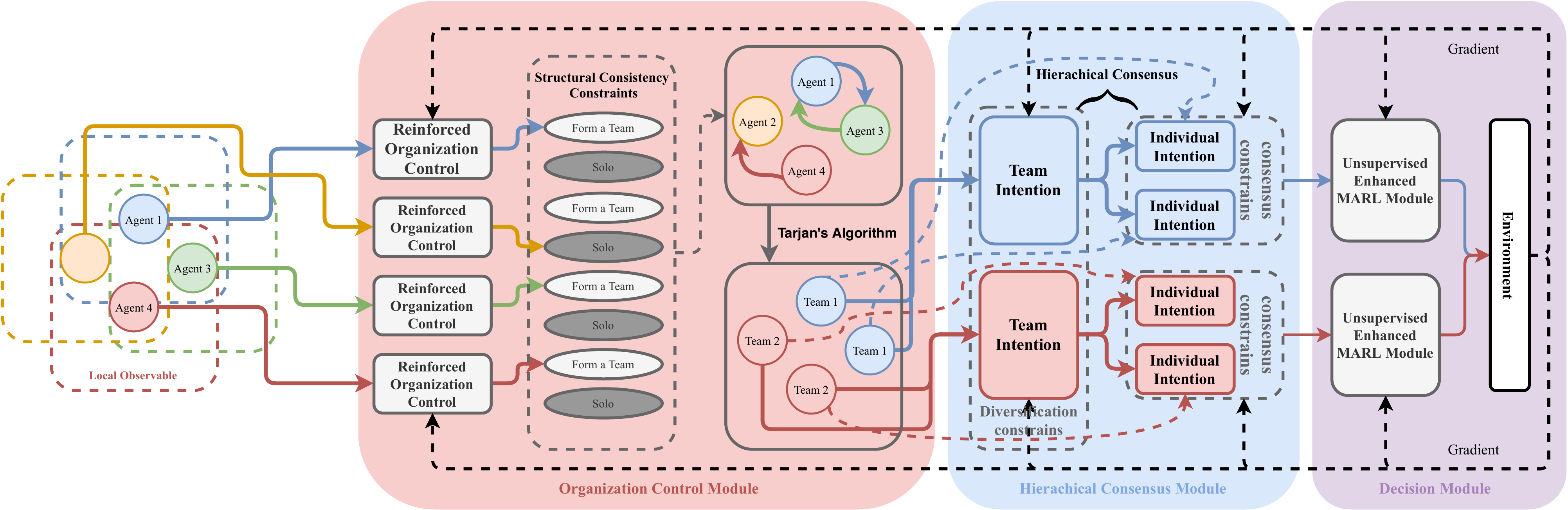}
  \caption{The {\sc{Rochico}} architecture. The algorithm is divided into three modules: organization control, hierarchical consensus, and decision. Each agent first makes a decision whether to team up with neighbors through the organization control module based on local observations. Next, the team intention generator and individual intention generator, trained based on the hierarchical consensus constraints, generate team intentions and individual intentions according to the teaming results. Finally, the decision-making module generates structured diversify policies based on team intentions and individual intentions.}
  \label{fig:arch}
\end{figure*}

The core assumption is that the behavior of an agent is determined by both the team goal and its perception of the environment, while diverse team goals and consistent environmental perception can lead to structured diversification emergence.
Designing an efficient algorithm for the agent to learn structured diversification emergence ability,
we need to answer the following important questions: 
1) \textbf{Adaptive organization control}: how do the agents form teams spontaneously, and the team composition can dynamically change to adapt to the external environment?
2) \textbf{Structured diversification}: How to maintain diversity in the behavior of different teams and to form tight cooperation between the agents within the team? 
3) \textbf{Behavior emergence}: How to combine the above two processes organically?
Let us conduct an in-depth analysis.

{\em{Adaptive organization control}}.
In a cooperative task, before taking actions, an individual will first assess whether her ability is sufficient to complete the task based on her observations of the environment.
If not, she will seek help from other individuals.
Besides, since the environment is locally observable, a team would be more powerful by fetching the information from multiple individuals.
Therefore, some existing works introduce teams in the MARL algorithms. 
However, these teams are constructed either through heuristics~\citep{zhang2009integrating,zhang2010self} or through end-to-end blackbox optimization~\citep{jiang2018learning,de2019multi,sheng2020learning}.
Compared with these methods to passively put individuals into the team, a more reasonable way is to let the individuals actively decide who to team up with.


{\em{Structured diversification}}.
After the teams are formed, the most direct way to measure the diversity of team behavior (or the goal) is to compare the differences in policies between all agents in different teams.
However, 
the number of agents in different teams are various, and the agent’s policy is usually a stochastic conditional probability distribution.
As a result, for large state-action space, directly comparing the difference among these conditional distributions (or agent's policies) is intractable.
From another perspective, the behavior of agents in the same team may be generated by a potential team intention.
If the intentions of different teams could be mapped into the same latent space, these intentions can be easily compared.
Within each team, the team intention guides the behaviors of agents, and each agent's perception of the surrounding environment can also influence agents' behaviors.
We can make a reasonable assumption: 
if two agents both have the same perception of the environment and the same goal, their behaviors should be closely coordinated.
At the same time, an effective perception should be able to reconstruct the surrounding environment.
This motivates us to impose consensus constraints on the agents' perception of the environment, thus we can achieve tight collaboration within the team.
A Similar idea can be found in \citet{mao2019neighborhood},
but it only imposes perceptual consensus constraints on the agent and other agents within its fixed neighborhood to encourage collaboration, which can't leads to agents' diverse behaviors.

In this paper, we propose a structured diversification emergence MARL algorithm, so-called {\sc{Rochico}}, based on reinforced organization control and hierarchical consensus learning.
As shown in Figure~\ref{fig:arch}, {\sc{Rochico}} consists of three modules: organization control, hierarchical consensus and decision.
First, the organization control module models the multi-agent organization control problem as a partially observable stochastic game (POSG), and introduces independent MARL to obtain a dynamic and autonomous teaming strategy.
Second, the hierarchical consensus module obtains team intentions and individual intentions through contrastive learning and unsupervised learning.
Structured diversification emerges by imposing hierarchical consensus constraints on hierarchical intentions.
Finally, the decision module outputs the diversify cooperative policies based on results of reinforced organization control and hierarchical consensus learning to the environment, and feeds back external rewards to the other two modules, and decision module itself, so that all modules can be combined organically.

Our contributions mainly consist of the following folds: 
1) We model the organization control of the multi-agent as an independent learning task, which enables the agent to autonomously and adaptively team up based on environmental feedback.
2) We impose hierarchical consensus constraints on hierarchical intentions obtained by the novel introduced contrastive learning and unsupervised learning auxiliary tasks to encourage the structured diversification emergence.
3) Performance experiments on various large-scale cooperative tasks show that {\sc{Rochico}} is significantly better than the current SOTA algorithms in terms of exploration and cooperation efficiency.


\section{Related Works}
\subsection{Organization Control Mechanism}
Organization control
is defined as a mechanism or a process that enables a system to change its organization without explicit command during execution time~\cite{di2005self}.
The most relevant existing self-organization mechanisms to our work can be summarized as \textit{task allocation}~\citep{macarthur2011distributed,ramchurn2010decentralized,dos2012distributed}, \textit{relation adaption}~\citep{gaston2005agent,Glinton2008AgentON,Kota2012DecentralizedAF} and \textit{coalition formation}~\citep{Chalkiadakis2010CooperativeGW,Mares2000FuzzyCS,Ye2013SelfAdaptationBasedDC}.
For the reason that these methods are not RL-based and we will not expand here, more details can be found in~\citet{ye2016survey}.
The task allocation refers to the agent actively allocates the task(s) to other agents because it cannot finish it by itself, which is different from classical task allocation in RL.
Further, the key difference between our work with relation adaption or coalition formation methods, is that our organization control mechanism is obtained through independent learning, rather than based on heuristic techniques or on large communication driven negotiation.

In recent years, there are a few works introduce organization control into reinforcement learning.
\citet{Abdallah2007MultiagentRL} uses RL to design the task allocation mechanism and transfer the learned knowledge across the different steps of organization control with heuristics mechanism.
\citet{zhang2009integrating, zhang2010self} integrate organization control into MARL to improve the convergence speed, which suffers large communication overhead because of the negotiation.


\subsection{Behavior Diversification}
Many cooperative tasks require agents to take different behaviors to achieve higher degree of task completion.
Behavior diversification can be handcrafted or emerged through multi-agent interaction.
Handcrafted diversification is widely studied as task allocation
or role assignment.
Heutistics mechanisums~\citep{sander2002scalable,dastani2003role,sims2008automated,macarthur2011distributed} assign a specific task or a pre-defined role to each agent based on its goal (capability, visibility) or by searching.
\citet{shu2018m} establishes a manager to assign suitable sub-tasks to rule-based workers with different preferences and skills.
All these methods require that the sub-tasks and roles are pre-defined, while the worker agents are rule-based at the same time.

Recently, the emergent diversification was introduced to single-agent RL~\citep{haarnoja2017reinforcement,haarnoja2018composable,haarnoja2018soft,eysenbach2018diversity} with the purpose to learn reusable diverse skills in complex and transferable tasks.
In MARL, \citet{mckee2020social} introduces diversity into heterogeneous agents to learn more generalized policies for solving social dilemmas.
\citet{wang2020multi} learns a role embedding encoder and a role decoder simultaneously.
However, no mechanism guarantees the role decoder can generate different parameters and generate diversity policies accordingly, taking as input different role embeddings.
\citet{jiang2020emergence} establishes an intrinsic reward for each agent through a well-trained probabilistic classifier.
The intrinsic reward makes the agents more identifiable and promote diversification emergence.
Based on \citet{eysenbach2018diversity}, learning low-level skills for each agent in hierarchical MARL is considered in~\citet{lee2019learning,yang2020hierarchical}.
The high-level policy can utilize coordinated low-level diverse skills, but the high-level policy does not consider diversity.

\section{Preliminaries}

\noindent\textbf{Cooperative POSGs.}
POSG~\citep{hansen2004dynamic} is denoted as a seven-tuple based on the stochastic game (or Markov game) as
$$
\langle \mathcal{X}, \mathcal{S}, \left\{ \mathcal{A}^i \right\}_{i=1}^{n}, \left\{ \mathcal{O}^i \right\}_{i=1}^{n}, \mathcal{P}, \mathcal{E}, \left\{ \mathcal{R}^i \right\}_{i=1}^{n} \rangle,
$$where $n$ denotes agents total number;
$\mathcal{X}$ represents the agent space;
$\mathcal{S}$ contains a finite set of states;
$\mathcal{A}^i$, $\mathcal{O}^i$ and  denote a finite action set and a finite observation set of agent $i$ respectively; $\boldsymbol{\mathcal{A}}=\mathcal{A}^1\times\mathcal{A}^2\times\cdots\times\mathcal{A}^n$ is the finite set of joint actions;
$\mathcal{P}(s^{\prime}|s, \boldsymbol{a})$ denotes the Markovian state transition probability function;
$\boldsymbol{\mathcal{O}}=\mathcal{O}^1\times\mathcal{O}^2\times\cdots\times\mathcal{O}^n$ is the finite set of joint observations;
$\mathcal{E}(\boldsymbol{o}|s)$ is the Markovian observation emission probability function;
$\mathcal{R}^i: \mathcal{S}\times\boldsymbol{\mathcal{A}}\times\mathcal{S} \rightarrow {\mathcal{R}}$ denotes the reward function of agent $i$.
The game in POSG unfolds over a finite or infinite sequence of stages (or timesteps), where the number of stages is called \textit{horizon}. 
In this paper, we consider the finite horizon case.
The objective for each agent is to maximize the expected cumulative reward received during the game. 
For a cooperative POSG, we quote the definition in \citet{Song2020ArenaAG},
$$
\forall x \in \mathcal{X}, \forall x^{\prime} \in \mathcal{X} \backslash\{x\}, \forall \pi_{x} \in \Pi_{x}, \forall \pi_{x^{\prime}} \in \Pi_{x^{\prime}}, \frac{\partial \mathcal{R}^{x^{\prime}}}{\partial \mathcal{R}^{x}} \geqslant 0, \nonumber
$$where $x$ and $x^\prime$ are a pair of agents in agent space $\mathcal{X}$;
$\pi_{x}$ and $\pi_{x^\prime}$ are the corresponding policies in the policy space $\Pi_{x}$ and $\Pi_{x^\prime}$ respectively.
Intuitively, this definition means that there is no conflict of interest for any pair of agents.


\noindent\textbf{QMIX.}
The QMIX~\citep{Rashid2018QMIXMV} algorithm, which is the follow-up work of the VDN~\citep{Sunehag2018ValueDecompositionNF}, is the current SOTA of cooperative MARL. 
QMIX claims that the the $Q$-value functions before ($Q_{t o t}$) and after ($Q_{a}$) decomposition should satisfy the following constraints: 
$$\frac{\partial Q_{t o t}}{\partial Q_{a}} \geq 0, \forall a \in A.$$
QMIX employs a hypernetwork-based mixing network to promote the two $Q$-function satisfying the above condition.
Because of the non-linear mixing network, QMIX can outperform VDN.

\section{Algorithms}
The overall framework of the proposed algorithm is shown in Figure~\ref{fig:arch}, which can be divided into three modules. 
The organization control module receives local observations of all agents and makes adaptive teaming decisions use the traditional graph theory algorithm~\cite{tarjan1972depth}.
The hierarchical consensus module will then generate the team intention and the individual intention based on the obtained teaming results.
The hierarchical consensus constraints are established on the above hierarchical intentions to promote the structured diversification emergence.
Finally, the decision module outputs the structured diversification policies through a cooperative MARL algorithm.
However, it should be noted that the organization control process is not differentiable, so we cannot train the overall model by the end-to-end scheme.
We draw on the communication MARL algorithms~\cite{foerster2016learning,kim2018learning,sheng2020learning}, while passing the external rewards to the organization control module and decision module separately, so as to realize the joint training.


\subsection{Organization Control Module}

If we consider all agents as a graph $\mathcal{G}({\mathcal{V}}, {\mathcal{E}})$, each agent is a node\footnote{Except for explicit emphasis, we will no longer distinguish between the two terms node and agent below.} $v \in {\mathcal{V}}$ in the graph ${\mathcal{G}}$.
The edge $e \in {\mathcal{E}}$ indicates whether the two agents connected by $e$ belong to the same team.
The main purpose of the organization control module is to determine the connections (edges) between agents.
Then, we can naturally view the connected components as teams,
while searching for connected components can be done efficiently by traditional graph theory algorithms~\cite{tarjan1972depth}.

In the beginning, if the problem in organization control module is modeled as a single-agent RL problem, $\mathcal{O}(n^2)$ edges need to be determined if the graph contains $n$ nodes,
and the size of the action space is $\mathcal{O}(2^{n^2})$.
This makes it impossible to scale to a larger multi-agent scenario (e.g., tens of agents).
Therefore, we model the organization control problem as a MARL problem, while each node is considered as an agent.
However, if each agent needs to determine its connections with all other nodes, the action space is still very large and be $\mathcal{O}(2^n)$.
Therefore, inspired by other team-based MARL algorithms, like \cite{jiang2018learning,jiang2019graph}, we only consider the closest $m$ other agents.
This idea is intuitive but effective, because it is reasonable to team up the agents according to the adjacency of the spatial position for most tasks.

Formally, the orgnization control problem can be modeled as a cooperative POSG, which is denoted as:
$$
\mathcal{M}_{u} := \left\langle \mathcal{X}_{u}, \mathcal{S}_{u}, \left\{ \mathcal{A}_{u}^i \right\}_{i=1}^{n}, \left\{ \mathcal{O}_{u}^i \right\}_{i=1}^{n}, \mathcal{P}_{u}, \mathcal{E}_{u}, \left\{ \mathcal{R}_{u}^i \right\}_{i=1}^{n} \right\rangle.
$$
We set the action of agent $i$, $a_{u}^i \in \mathcal{A}_{u}^i$, as a $m$-dimension binary vector, denoting the connection action to $m$-nearest agents $\{x^j \mid j \in \mathcal{N}_m(i)\}$ ($m\ll n$) according to its local observation $o_{u}^i$, where the subscript $u$ stands for 'unorganized'.
This kind of MARL problem will suffer that agent $i$ decides to connect with agent $j$ but agent $j$ does not want to connect with agent $i$ or vice versa, which means:
$$
a_{u}^i [j] \neq a_{u}^j [i],\quad {\rm with}\quad i \in \mathcal{N}_{m}(j),\;j \in \mathcal{N}_{m}(i).
$$
Considering the positive effect of prosociality on promoting the cooperation of agents~\cite{peysakhovich2018prosocial}, we use the weakly connected graph $\mathcal{G}_{u}= ({\mathcal{V}}_u, {\mathcal{E}}_u)$ to form teams, which is established by converting directed edges into undirected edges, i.e.,
$$
e(i,j) = a_{u}^i[j] \vee a_{u}^j[i],\quad {\rm with} \quad i \in \mathcal{N}_{m}(j),\;j \in \mathcal{N}_{m}(i),
$$where $\vee$ denotes ``{\sc{or}}" operation.
Finally, the Tarjan's algorithm is employed for searching weakly connected components with worst-case time complexity as $\mathcal{O}(|{\mathcal{V}}_u|+|{\mathcal{E}}_u|)$~\cite{tarjan1972depth}.

In addition to the external rewards $\{r_{e}^i\}_{i=1}^n$ from the environment, we also introduce additional intrinsic rewards to train the agents.
Specifically, in order to strengthen training stability without causing excessive fluctuation on the graph structure, the novel \textit{structural consistency intrinsic reward} $r_{u}^i$ for each agent $i$ is defined by:
$$
r_{u}^{i} = \frac{1}{|\mathcal{N}_{m}(i)|}\cdot {\rm GED} \left( \mathcal{G}_{u} \left( \mathcal{N}_{m}(i) \vee i \right), \; \mathcal{G}_{u}^{'} \left( \mathcal{N}_{m}(i) \vee i \right) \right),
$$where $\text{GED}(\cdot,\cdot)$ represents the \textit{graph edit distance}~\citep{sanfeliu1983distance}.
$\mathcal{G}_{u}\left(\mathcal{N}_{m}(i)\vee i\right)$ and $\mathcal{G}_{u}^{'}\left(\mathcal{N}_{m}(i)\vee i\right)$ represent the sub-graph only contains node $i$ and its $m$-nearest neighbors before and after take action $a_{u}^i$ respectively.
The total reward for each agent $i$ is:
$$
r_{u+}^{i} = r_{e}^{i} + \alpha_{u} r_{u}^{i},
$$where $\alpha_u$ indicates the strength of contraint for structural consistency.
For the organization control problem, the goal is to maximize the summation of all agents' expected accumulated rewards, which can be denoted as follows:
$$
\max\ \mathcal{J}_{u} = \frac{1}{n}\sum_{i=1}^{n}\mathbb{E}_{\tau_u}\left[ r_{u+}^i(\tau_u) \right].
$$
Further to raise training efficiency, we use independent learning combined with DQN and parameter sharing mechanism.
\cite{papoudakis2020comparative} has shown the significant performance of independent learning on multi-agent tasks.
Specifically, we can minimize following TD($0$) error for each agent $i$, i.e.,
$$
\mathcal{L}_{u}^i(\theta_u) = \mathbb{E}_{\left(o_u^i,a_u^i,r_{u+}^i,o_u^{i,'}\right)\sim D}\left[ \left(Q_{\theta_u}(o_u^i,a_u^i)-y\right)^2 \right].
$$where $y = r_{u+}^i + \gamma \max_{a_u^{i,'}} Q_{\bar{\theta_u}}(o_u^{i,'},a_u^{i,'})$ and $\theta_u, \bar{\theta_u}$ parameterize $Q$ function and target $Q$ function separately.
Finally, the overall objective function is:
$$
\min \mathcal{L}_{u}^{Q}(\theta_u) = \sum_{i} L_{u}^i(\theta_u).
$$

\subsection{Hierarchical Consensus Module}
The goal of the hierarchical consensus module is to achieve efficient multi-agent exploration and cooperation, that is, structured diversification emergence.
To this end, we put forward the concepts of hierarchical intentions, i.e., team intentions and individual intentions. 
Then the structured diversification emergence can be achieved through contrastive learning and hierarchical consensus learning.
The hierarchical consensus module is composed of two sub-modules: team intention generator (Figure~\ref{fig:gcg}) and individual intention generator (Figure~\ref{fig:icg}).


\subsubsection{Team Intention Generator}

\begin{figure}[htbp]
  \centering
  \includegraphics[width=0.5\textwidth]{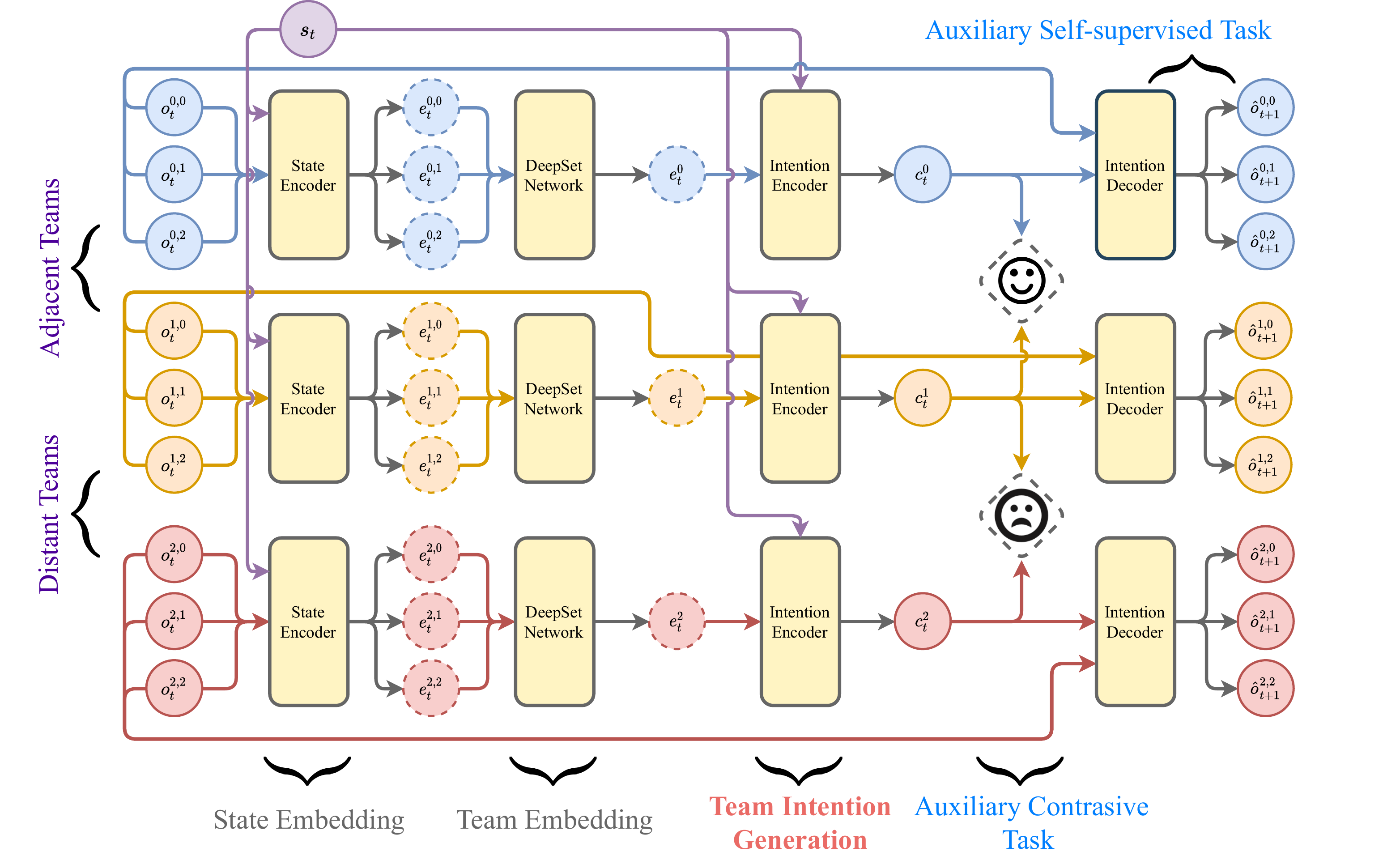}
  \caption{The network structure of the team intention generator in the training phase. Two teams are close to each other and one team is far apart. 
  The team intention is based on the representations of all agents in the team, which are aggregated through a DeepSet network, and further is calculated through the intention encoder. The loss function consists of both contrastive loss and self-supervised loss.}
  \label{fig:gcg}
\end{figure}

The team intention should have the following characteristics:
1) the team intention must reflect the team behavior (or goal), so that it should be generated based on the joint observation of all agents within the team;
2) to improve the exploration efficiency of the team in an unknown environment, the team intentions among different teams must be diverse;
3) the team intention can reflect the team behaviors in a short period time, so we can predict the future observation of the agents within the team based on the team intention. 
In addition to the second one, the agents in each team make decisions independently, so they must access the local information of the other agents in the same team to achieve diverse behaviors.
Considering the difficulty of communication learning~\cite{lowe2019pitfalls}, we use the global state of environment additionally to assist the generation of team intention, similar as~\citet{rashid2018qmix}.
The global state is only used in the training phase since we use a CTDE framework.

Formally, in order to generate the team intention for each team, the state encoder $f_{\mu}(\cdot)$ parameterized by $\mu$ recieves the joint observation $\boldsymbol{o}_{t}^{k}=(o_{t}^{k,1}, o_{t}^{k,2}, \cdots, o_{t}^{k, n_k})$ of all $n_k$ agents in team $k$, and together with the global state $s_t$ (such as the minimap of the environment) at timestep $t$.
The generated agent state embeddings of each team $(e_{t}^{k,1}, e_{t}^{k,2}, \cdots, e_{t}^{k, n_k})$ are feed into a DeepSet network~\cite{zaheer2017deep}, i.e., $f_{\nu}(\cdot)$ parameterized by $\nu$ to generate the team embedding ${e}_t^{k}$:
$$
{e}_t^{k} = f_{\nu}\left( f_{\mu}\left(o_{t}^{k,1}\right), f_{\mu}\left(o_{t}^{k,2}\right), \cdots, f_{\mu}\left(o_{t}^{k,n_k}\right) \right).
$$Finally, the team intention encoder $f_{\omega}(\cdot)$ parameterized by $\omega$ recieves team embedding ${e}_t^{k}$ and global state $s_t$ again to generate the team intention ${c}_t^{k}$:
$$
{c}_t^{k} = f_{\omega}\left( {e}_t^{k}, s_t \right).
$$

In order to generate diverse team intentions, we model the team intention generation as a \textit{contrastive learning} problem. 
First, the average spatial position $[\bar{x}_t^k, \bar{y}_t^k]$ of the team $k$ at each timestep $t$ can be calculated based on the spatial position of the all agents within the team $\{(x_t^{k,1}, y_t^{k,1}),\cdots,(x_t^{k,n_k}, y_t^{k,n_k})\}$.
Combined with the current timestamp, $t_k$, a spatiotemporal feature representation $e_{ts}^{k}=[\bar{x}_t^k, \bar{y}_t^k, t_k]$ of the team $k$ can be obtained.
The Euclidean distance in the spatiotemporal space is used to measure the distance between team $k$ and $l$, i.e.,  
$$
d(k, l) = \| e_{ts}^k - e_{ts}^l \|_2^2.
$$
A simple version directed graph $\mathcal{G}_{\ell}({\mathcal{V}}_{\ell}, {\mathcal{E}}_{\ell})$ can be established in team level,
where ${\mathcal{V}}_{\ell}$ denotes the node set of $\mathcal{G}_{\ell}$ and each node $v^k \in {\mathcal{V}}_{\ell}$ represents a team $k$.
The ${\mathcal{E}}_{\ell}$ denotes the edge set and the two teams connected by edge $e^k \in {\mathcal{E}}_{\ell}$ have similar intentions.
For each edge in edge set ${\mathcal{E}}_{\ell}$ of $\mathcal{G}_{l}$, we have
$$
e(k,l) := \left\{ \mathbf{1}[d(k, l)=\min_{v} d(k, v)] \right\} \vee \left\{ \mathbf{1}[d(l, k)=\min_{v} d(l, v)] \right\}.
$$
There is an edge connection between $k,l$ two nodes iff anyone is the nearest neighbor of the other.
Similar to the organization control module, we use Tarjan's algorithm to find all the weakly connected components in the directed graph ${\mathcal{G}}_{\ell}$.
Teams that belong to the same weakly connected component will be assigned the same label.
The team intention $c_{t}^{k}$ is set to be the feature and the Tarhan's algorithm label $y_{t}^k$ is set to be the label, which 
constructs a supervised training set $\left\{(c_{t}^{k}, y_{t}^k) \right\}$. 
After combining with the triplet loss~\cite{schroff2015facenet}, we could construct a contrastive learning problem by minimize following objective for each team $k$:
$$
\mathcal{L}^{k}_\ell(\mu,\nu,\omega)=\mathbb{E}\left[\max \left(0,\left\|c_{t}^{k} - c_{t}^{u}\right\|_{2}^{2}-\left\|c_{t}^{k} - c_{t}^{v}\right\|_{2}^{2}+m\right)\right],
$$ where $m$ is a margin parameter, $k,u$ share the same label ($y_t^k=y_t^u$) and $k,v$ have different labels ($y_t^k \neq y_t^v$).
$u$ and $v$ are sampled from the weak connected component same as $k$ and different from $k$ respectively.

In addition, in order to enhance the impacts of team intention on the team behavior, besides using the team intention as the input of the following individual intention generator, we introduce another self-supervised task.
The intention decoder $f_{\xi}(\cdot)$ parameterized by $\xi$ recieves 
team intention $c_t^k$ of team $k$ at current timestep $t$ as input, and output the prediction of joint observation $\boldsymbol{\hat{o}}_{t+1}^{k}=(\hat{o}_{t+1}^{k,1}, \hat{o}_{t+1}^{k,2}, \cdots, \hat{o}_{t+1}^{k, n_k})$ at next timestep.
Formally, we minimize following regression objective function for each team $k$ so as to formulate the self-supervised task, i.e.,
$$
\mathcal{L}^k_{\ell}(\xi) = \mathbb{E}\left[\frac{1}{n_k}\sum_{i=1}^{n_k} \left\| f_{\xi}(o_{t}^{k,i}, c_t^k) - o_{t+1}^{k,i} \right\|_2^2\right].
$$
The network structure is shown in Figure~\ref{fig:gcg} and the overall problem of the team intention generator can be formulated as follows
$$
\min\mathcal{L}_{\ell}^{tg}(\mu,\nu,\omega,\xi) = \mathbb{E}\left[\sum_{k} \mathcal{L}^{k}_{\ell}(\mu,\nu,\omega) + \lambda_{tg}\cdot\mathcal{L}^k_{\ell}(\xi)\right].
$$


\subsubsection{Individual Intention Generator}

\begin{figure*}[htbp]
  \centering
  \includegraphics[width=0.8\textwidth]{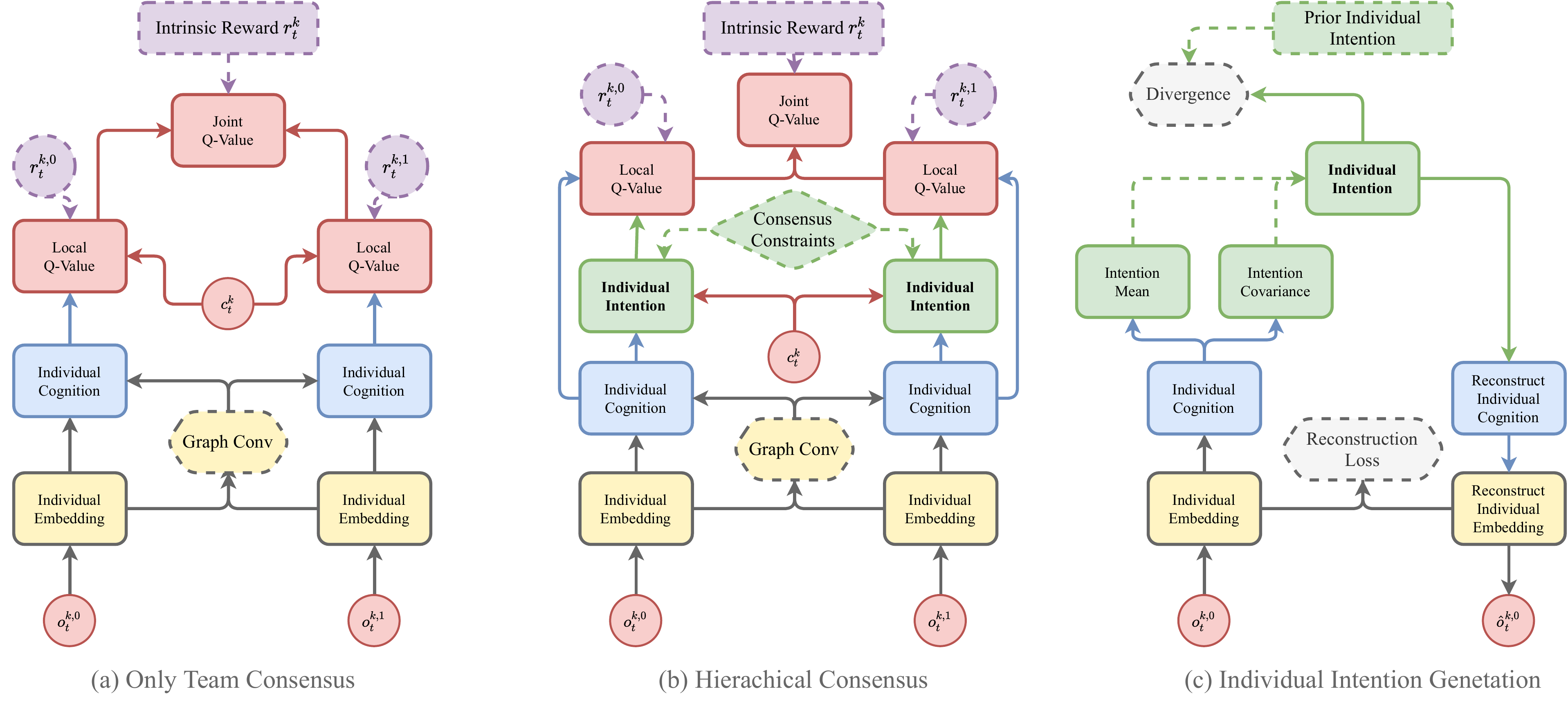}
  \caption{The network structure of the hierarchical consensus module and decision module. (a) The baseline structure that only uses team intention to generate final policies. (b) The individual intention is generated based on the team intention and consensus constraints. (c) The individual intention generator of {\sc{Rochico}} algorithm is a standard variational autoencoder.}
  \label{fig:icg}
\end{figure*}

With the team intention as the guidance for the diverse behavior between teams, the hierarchical consensus module aims to realize the structured diversification emergence through individual intentions consensus.
The key idea to achieve better cooperation within the team is that all agents in the same team should have a consistent cognition of the surrounding environment and the task.
The individual intention generation is divided into the following three steps:\\
\noindent{\bf{1)}}. The individual encoder $g_{\phi}$ parameterized by $\phi$ encodes the local observation $o_t^{k,i}$ (of agent $i$ in team $k$ at timestep $t$) into individual embedding $h_{t}^{k,i}$ (yellow rounded square in Figure~\ref{fig:icg}(b)), which only contains agent-specific information;

\noindent{\bf{2)}}. We regard agents in the same team as nodes in a new fully-connected graph.
A GNN $g_{\psi}$ parameterized by $\psi$ then is introduced to further aggregate all individual embeddings ${\boldsymbol{h}}_{t}^k=(h_{t}^{k,1},\cdots,h_{t}^{k,n_k})$ and extracts the individual cognition $\chi_{t}^{k,i}$ (blue rounded square in Figure~\ref{fig:icg}(b)) by
$$
\chi_{t}^{k,i} = g_{\psi}\left(\textstyle{\sum}_{j\in k}h_{t}^{k,j}\right);
$$

\noindent{\bf{3)}}. The variational encoder $g_{\varphi}$ (shown in Figure~\ref{fig:icg}(c)) recieves the concatenated feature $(\chi_{t}^{k,i},c_{t}^{k})$ (consist of individual cognition $\chi_{t}^{k,i}$ and team intention $c_{t}^{k}$ of team $k$) as input, and output the individual intention $\zeta_{t}^{k,i}$ (green rounded square in Figure~\ref{fig:icg}(b)) through\footnote{Since the reparameterization trick is used in order to enable the backpropagation during implementation, it is reasonable to write the equal sign here.}
$$
\zeta_{t}^{k,i} = g_{\varphi}(\chi_{t}^{k,i},c_{t}^{k}).
$$
Then, we impose hierarchical consensus constraints on the generated individual intentions, which leads to optimizing the following function for each agent $i$ in team $k$
$$
\mathcal{L}^i_{\ell}(\phi,\psi,\varphi)=\mathbb{E}\left[\frac{1}{n_k{-}1}\sum_{j \neq i} \text{KL}\left[ q_{\varphi}(\zeta_{t}^{k,i}|o_t^{k,i}) \big\| q_{\varphi}(\zeta_{t}^{k,j}|o_t^{k,j}) \right]\right].
$$
Recall the loss function of the variational autoencoder
$$
\mathcal{L}^i_{\ell}(\varphi) = \mathbb{E}\left[\| o_t^{k,i} - \hat{o}_t^{k,i} \|_2^2 + \text{KL}\left[ q_{\varphi}(\zeta_{t}^{k,i}|o_t^{k,i}) \big\| p(\zeta_{t}^{k,i}) \right]\right].
$$The prior distribution $p(\zeta_{t}^{k,i})$ in second term can be replaced by our above consensus constraints, which leads to
$$
\begin{aligned}
&\mathcal{L}^i_{\ell}(\phi,\psi,\varphi)=\mathbb{E}\bigg[\| o_t^{k,i} - \hat{o}_t^{k,i} \|_2^2 + \\
&\quad\quad \frac{1}{n_k{-}1}\sum_{j \neq i} \text{KL}\left[ q_{\varphi}(\zeta_{t}^{k,i}|o_t^{k,i}) \big\| q_{\varphi}(\zeta_{t}^{k,j}|o_t^{k,j}) \right]\bigg].
\end{aligned}
$$
The overall problem of individual intention generation can be formulated as
$$
\min \mathcal{L}_{\ell}^{ig}(\phi,\psi,\varphi) := \sum_{i=1}^n \mathcal{L}^i_{\ell}(\phi,\psi,\varphi).
$$


\subsection{Decision Module}
Before outputs the agents' policies, it should be noted that the team intention to affect the final policies indirectly.
To affect the agents' behaviors more directly and efficiently, we introduce another intrinsic team reward and utilize the QMIX~\cite{rashid2018qmix} technique to assign the rewards to all agents.
Formally, the intrinsic team reward $r_{t,\ell}^{k}$ is defined as:
$$
r_{t,\ell}^{k} = \sum_{j \neq k} \left\| c_t^k - c_t^j \right\|_2^2.
$$
We additionally use the external reward $r_{e}^i$ to learn the local $Q$ function, as a result the local $Q$ function will be trained based on two different reward signals $r_{e}^i$ and $r_{t,\ell}^{k}$.
To make the behavior of the agents in each team have a certain diversity, we firstly combine the individual intention $\zeta_{t}^{k,i}$ with the individual cognition $\chi_{t}^{k,i}$ as the input of the QMIX algorithm, and generates the local $Q$-value.
Further we optimize the local $Q$-value function by minimize following TD($0$) error:
$$
\mathcal{L}_{\ell}^i(\theta^i_\ell) = \mathbb{E}_{\left(\zeta_{t}^{k,i}, \chi_{t}^{k,i}, a_{t,\ell}^i\right)\sim D}\left[ \left(Q_{\theta^i_\ell}(\zeta_{t}^{k,i}, \chi_{t}^{k,i}, a_{t,\ell}^i)-y^i\right)^2 \right].
$$where $y^i = r_{\ell}^i + \gamma \max_{a_{t+1,\ell}^{i}} Q_{\bar{\theta_\ell^i}}(\zeta_{t+1}^{k,i}, \chi_{t+1}^{k,i}, a_{t+1,\ell}^i)$ and $\theta_{\ell}^i, \bar{\theta_{\ell}^i}$ parameterize local $Q$ function and local target $Q$ function of agent $i$ respectively.
For each team $k$, the team joint $Q$-value can be denoted as
$$
\begin{aligned}
Q_{\theta^k_\ell}(\boldsymbol{\zeta}_{t}^{k}, \boldsymbol{\chi}_{t}^{k}, \boldsymbol{a}_{t,\ell}^{k}) = Q_{\theta^k_\ell}\bigg(&Q_{\theta^1_\ell}(\zeta_{t}^{k,1}, \chi_{t}^{k,1}, a_{t,\ell}^1), \\
&\cdots, Q_{\theta^{n_k}_\ell}(\zeta_{t}^{k,n_k}, \chi_{t}^{k,n_k}, a_{t,\ell}^{n_k})\bigg),
\end{aligned}
$$and we also optimize the team joint $Q$-value function by minimize following TD($0$) error:
$$
\mathcal{L}_{\ell}^k(\theta^k_\ell) = \mathbb{E}_{\left(\boldsymbol{\zeta}_{t}^{k}, \boldsymbol{\chi}_{t}^{k}, \boldsymbol{a}_{t,\ell}^{k}\right)\sim D}\left[ \left(Q_{\theta^k_\ell}(\boldsymbol{\zeta}_{t}^{k}, \boldsymbol{\chi}_{t}^{k}, \boldsymbol{a}_{t,\ell}^{k})-y^k\right)^2 \right].
$$where $y^k = r_{t,\ell}^{k} + \gamma \max_{\boldsymbol{a}_{t+1,\ell}^{k}} Q_{\bar{\theta_\ell^k}}(\boldsymbol{\zeta}_{t+1}^{k}, \boldsymbol{\chi}_{t+1}^{k}, \boldsymbol{a}_{t+1,\ell}^{k})$ and and $\theta_{\ell}^k, \bar{\theta_{\ell}^k}$ parameterize joint $Q$-value function and joint target $Q$-value function of team $k$ respectively.
The overall learning problem of decision module is
$$
\min\mathcal{L}_{\ell}^{Q}(\{\theta_\ell^i\},\{\theta_\ell^k\}) = \sum_{i} \mathcal{L}_{\ell}^i(\theta^i_\ell) + \lambda_{QMIX} \sum_{k} \mathcal{L}_{\ell}^k(\theta^k_\ell).
$$


\section{Experiments}

\subsection{Environments}
We evaluate the algorithm performances on four large-scale cooperative environments, including {\em{Pacmen}}, {\em{Block}}, {\em{Pursuit}} and {\em{Battle}}.
More details can be found in appendix.

\noindent{\textbf{Pacmen}}. $64$ agents initialized at the maze center and $64$ dots scatter randomly at four corners of the squared map.
Agents get the reward by eat dots.
The dots are distributed in different corners, the agent needs to team up and travel to different corners to eat more dots.

\noindent{\textbf{Block}}. There are $32$ blockers and $32$ blockees who have superior speed than the blockers.
There also are $64$ foods initialized at one side of the squred map.
Blockers and blockees are only rewarded by eat foods.
Since blockee runs faster than blocker, blocker needs to learn diverse policies to block blockees and eat food Simultaneously.

\noindent{\textbf{Pursuit}}. There are $64$ predators and $64$ preys who have superior speed than the predators. 
Since the prey runs faster than the predators, the predators need to learn to round up through structured diversification policies.

\noindent{\textbf{Battle}}. $64$ agents learn to fight against $64$ enemies who have superior abilities than agents. 
As the hit point of enemy is $10$ (more than single agent's damage), agents need to continuously cooperate to kill the enemy.
All environments are implemented by MAgent~\citep{zheng2018magent}.

\subsection{Baselines}
IDQN is chosen as the baseline.
Due to the connection between {\sc{Rochico}} with QMIX and NCC-Q, we also compare these two methods as baselines.
However, both QMIX and NCC-Q are not designed for large-scale scenarios. 
Therefore, we first randomly split all agents to multiple teams, and use QMIX and NCC-Q algorithms in each one.
See Appendix for detailed hyperparameter settings.

\subsection{Performance Comparison}
\begin{figure*}
    \centering
    \subfigure[The performance comparison in Pacmen environment.]{
    \includegraphics[width=0.31\textwidth]{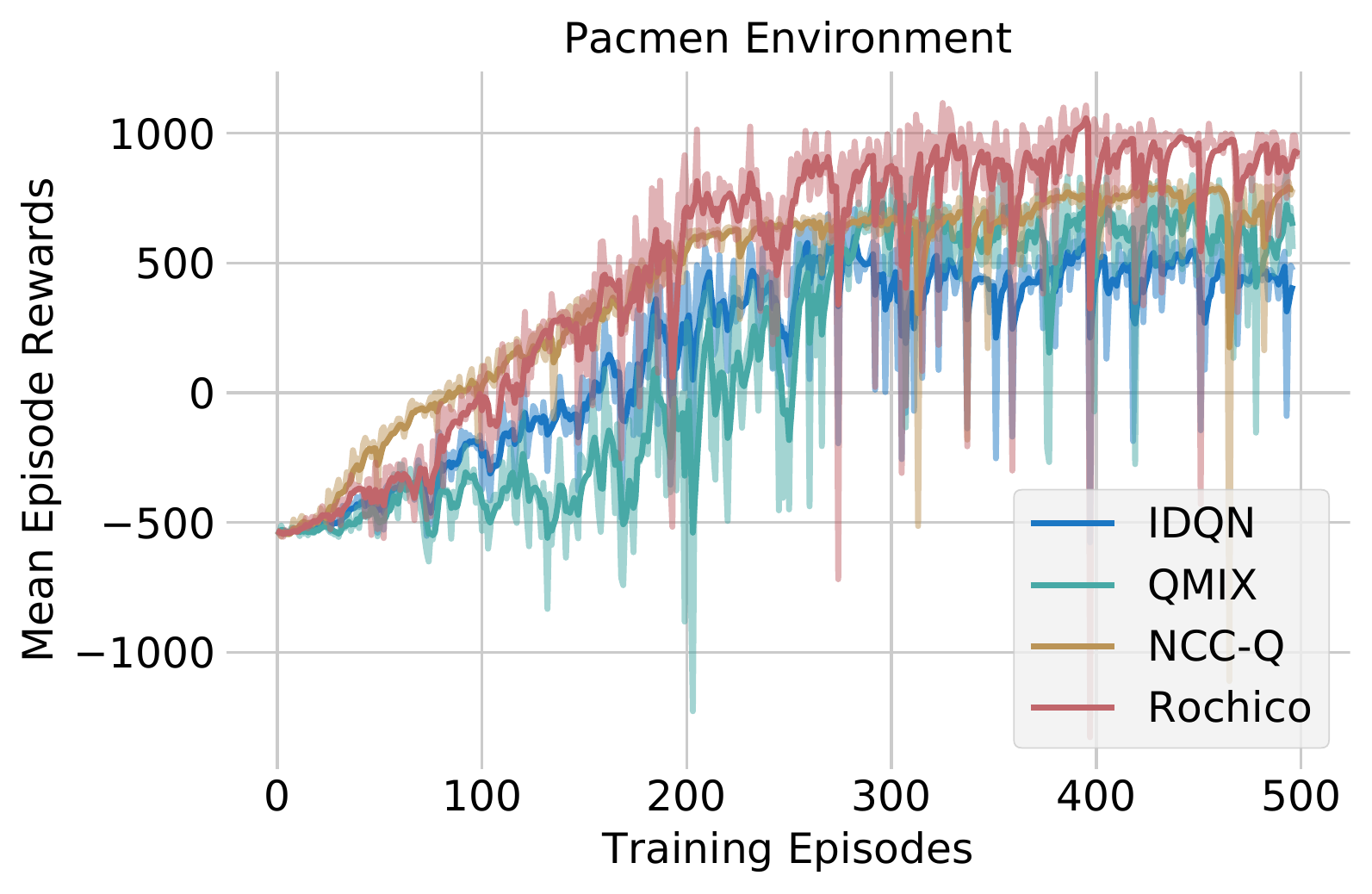}
    \label{fig:pacmen-totalreward}
    }
    \subfigure[The performance comparison in Blcok environment.]{
    \includegraphics[width=0.31\textwidth]{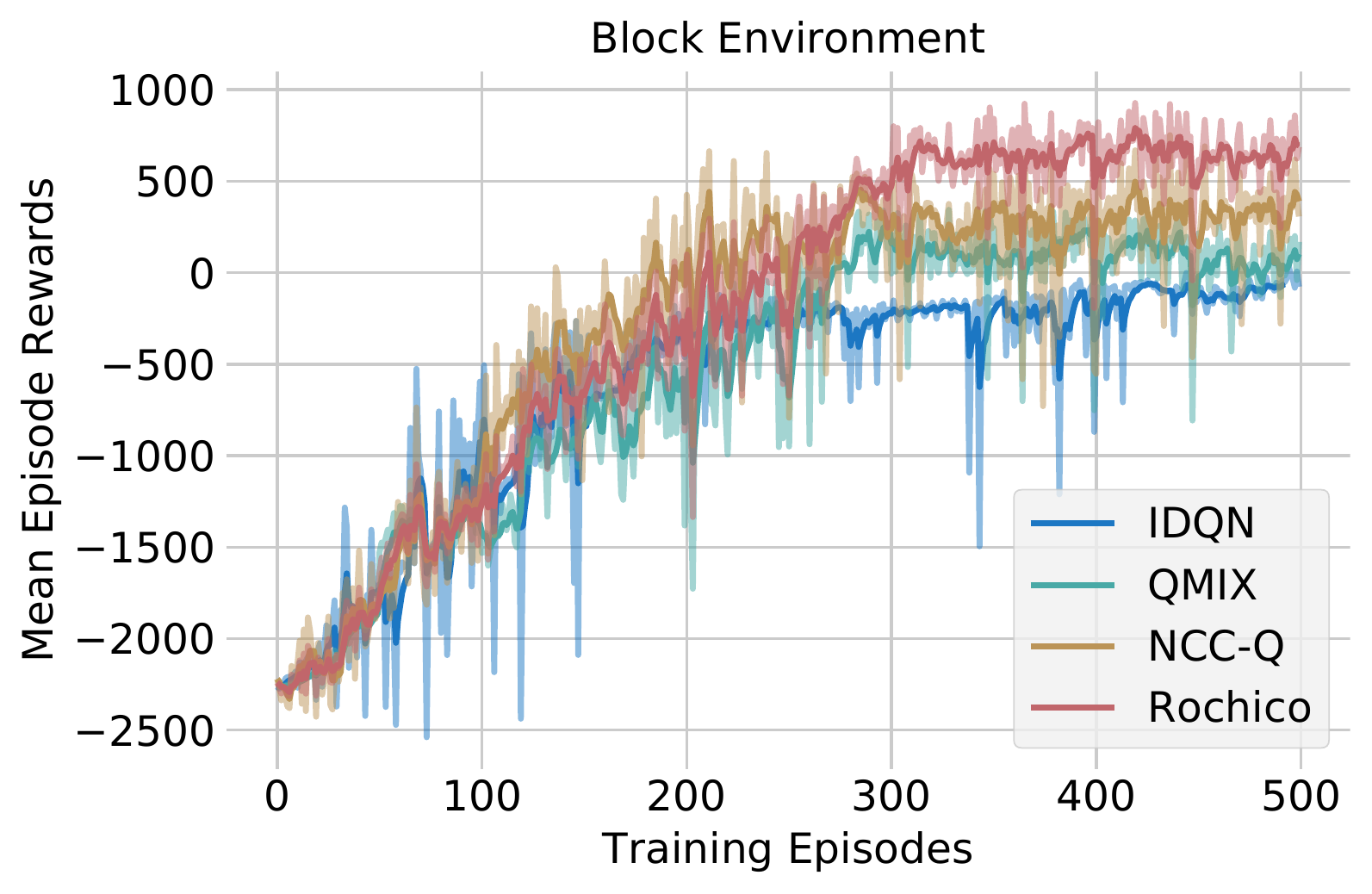}
    \label{fig:block-totalreward}
    }
    \subfigure[The performance comparison in Pursuit environment.]{
    \includegraphics[width=0.31\textwidth]{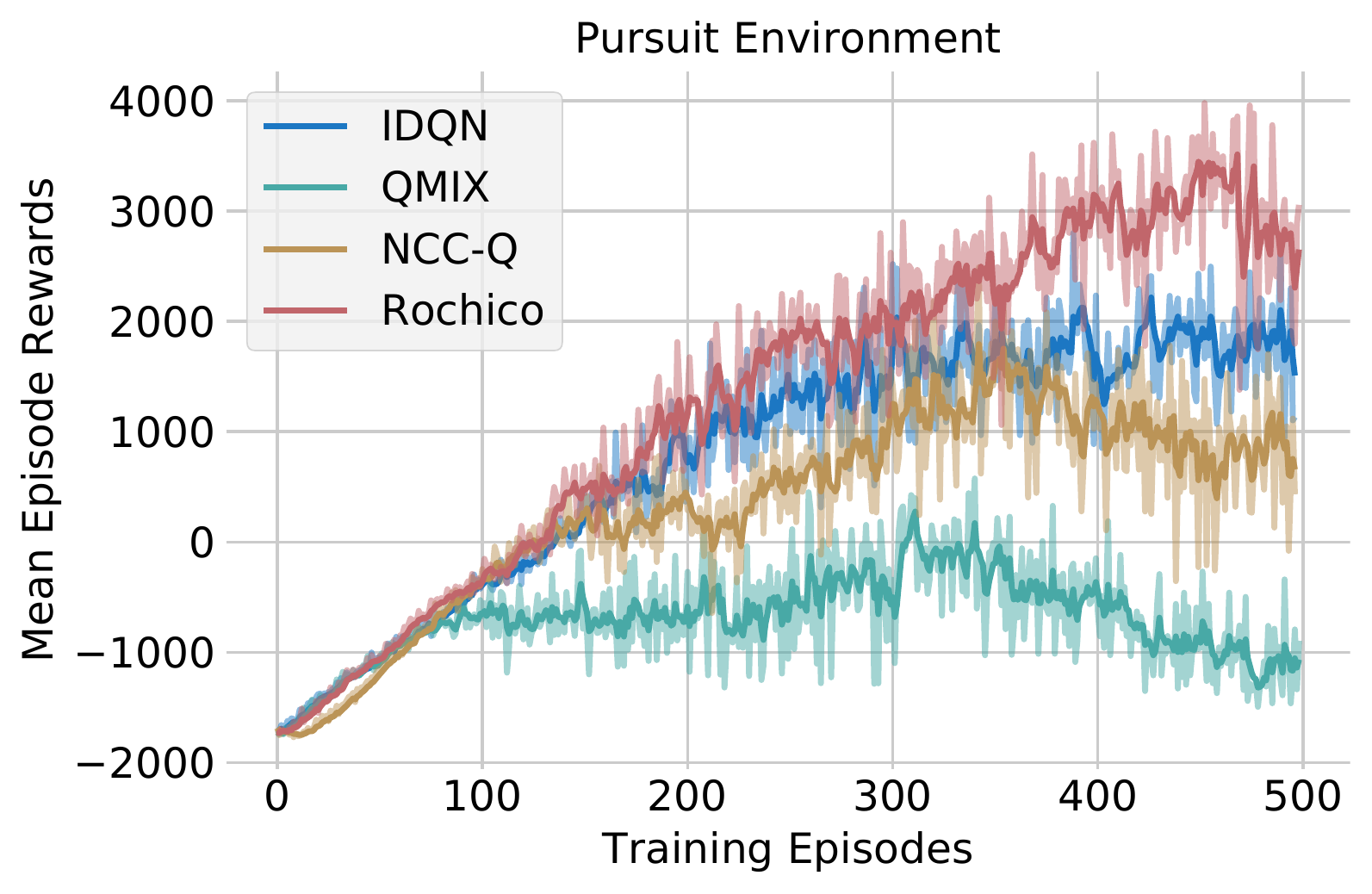}
    \label{fig:pursuit-totalreward}
    }
    \subfigure[The performance comparison in Battle environment.]{
    \includegraphics[width=0.31\textwidth]{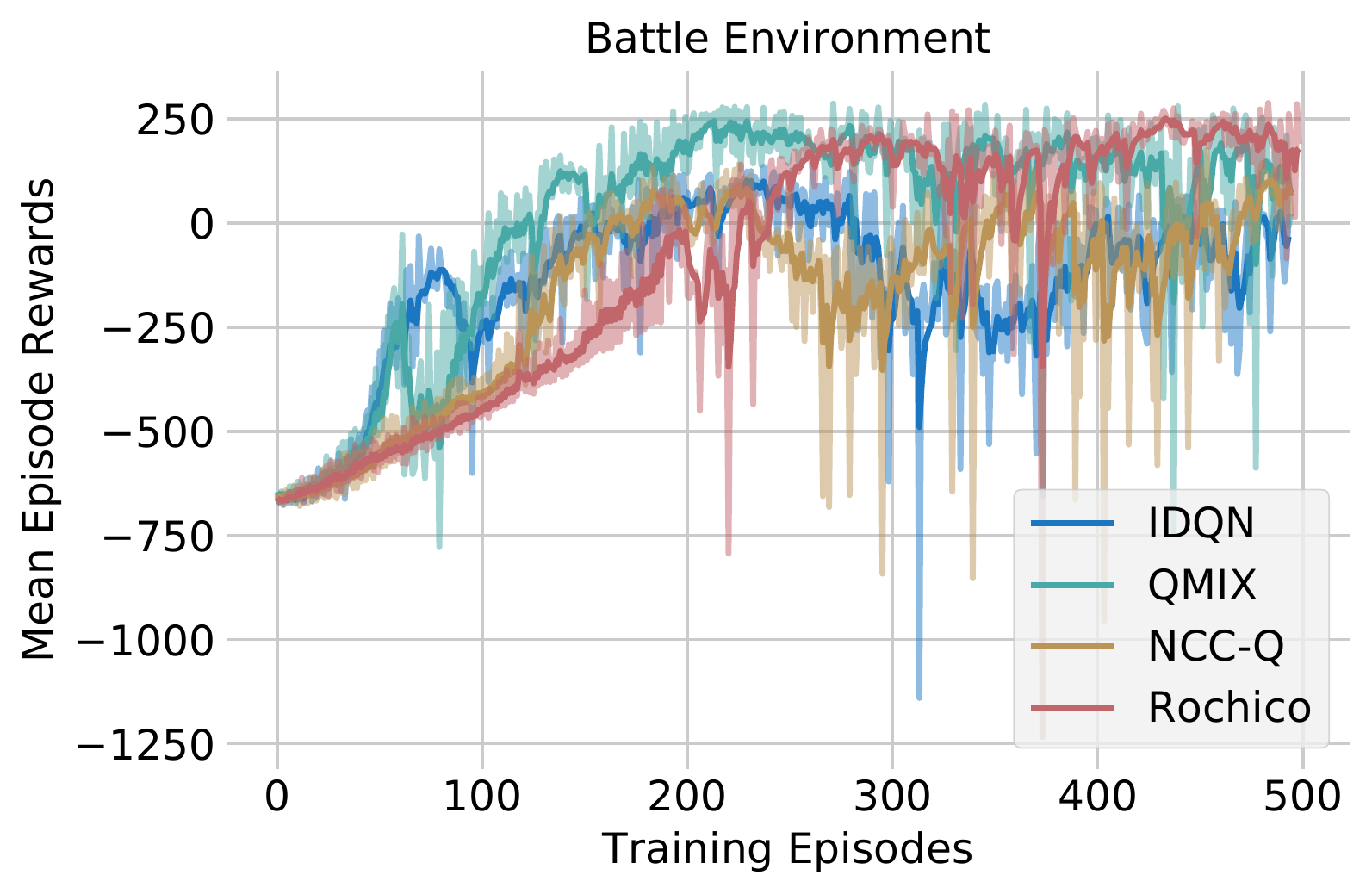}
    \label{fig:battle-totalreward}
    }
    \subfigure[The ablation study in Pursuit environment.]{
    \includegraphics[width=0.31\textwidth]{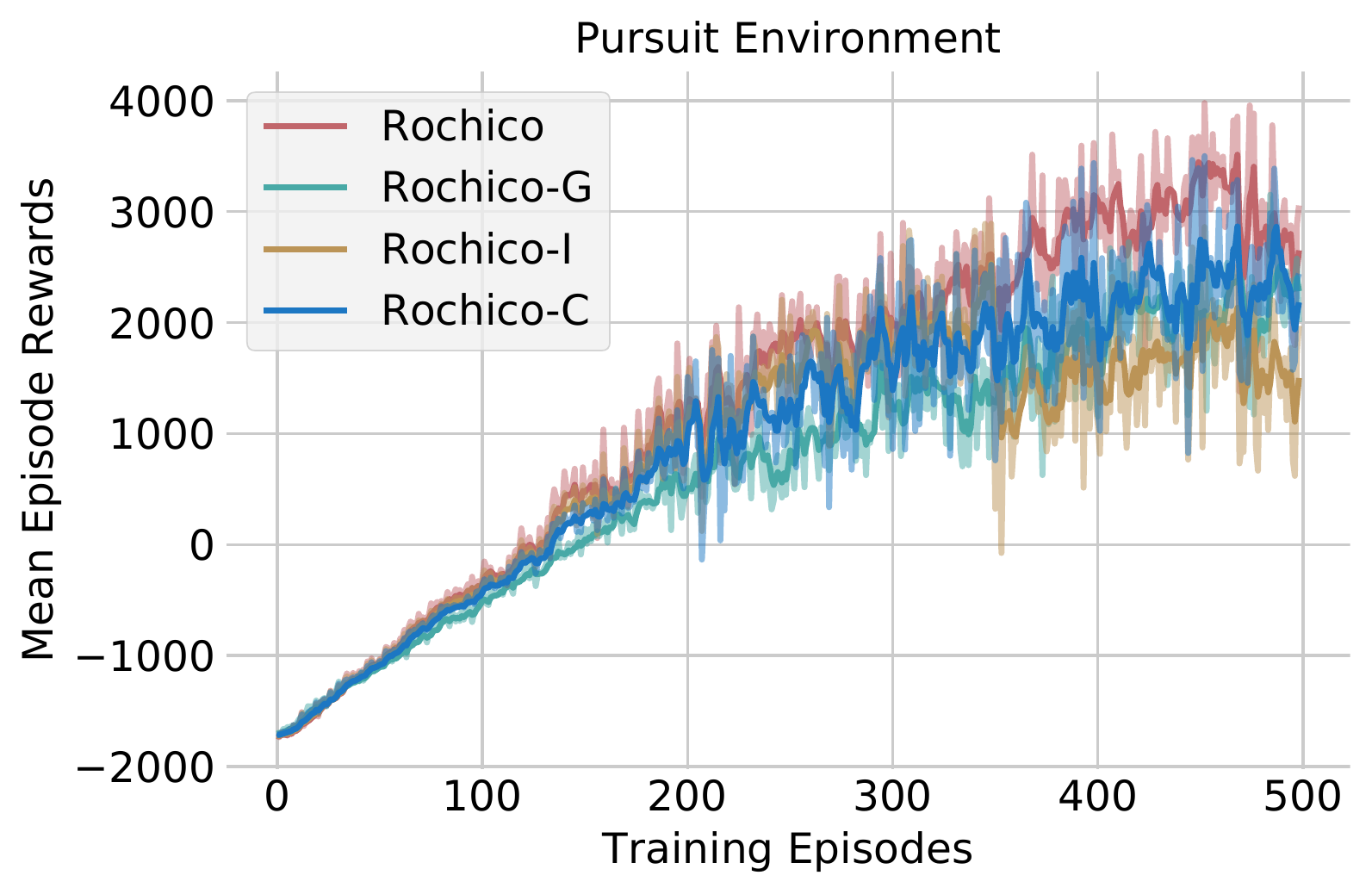}
    \label{fig:pursuit-ablation}
    }
    \subfigure[The ablation study in self-orgnization phase.]{
    \includegraphics[width=0.31\textwidth]{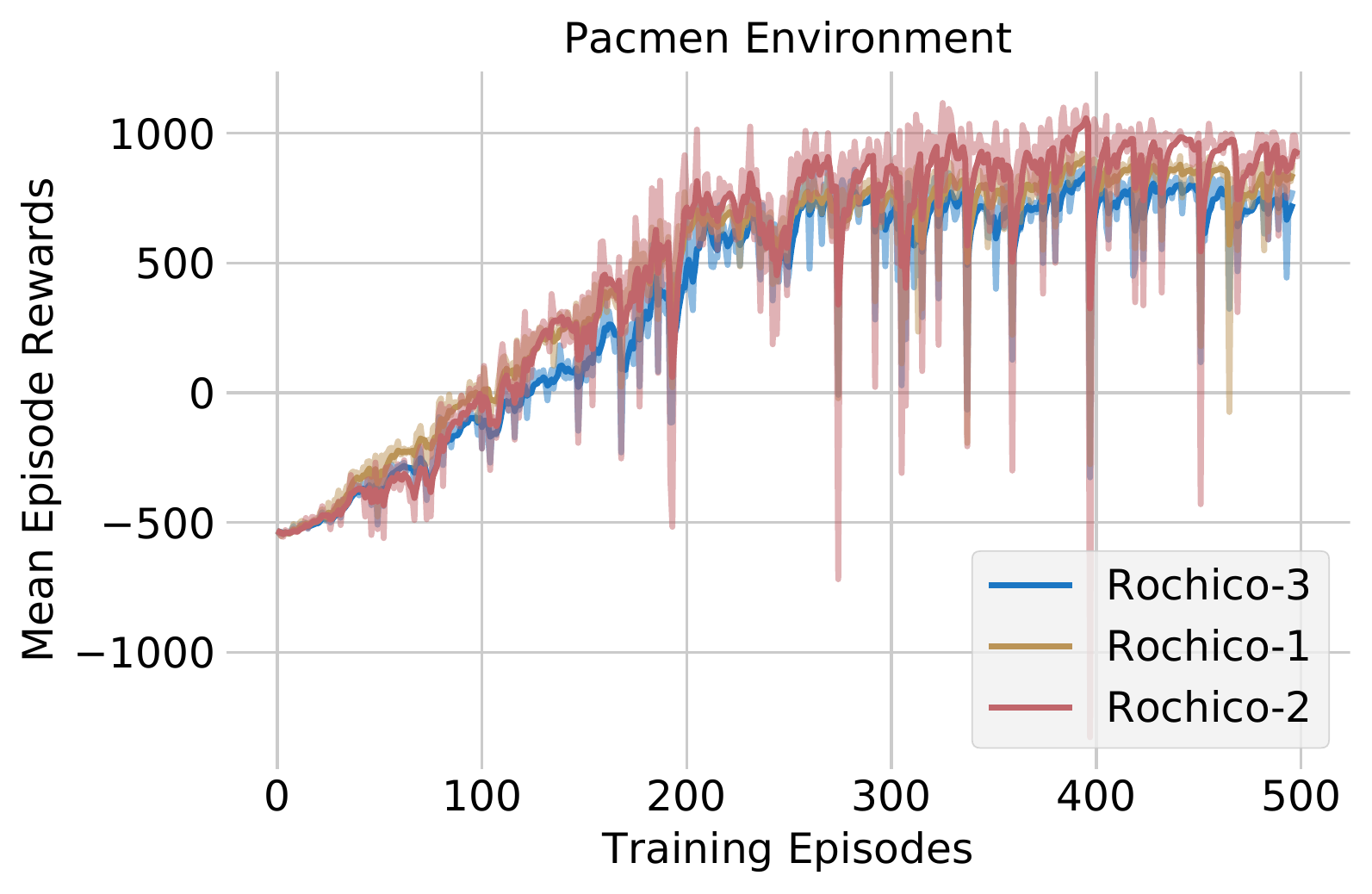}
    \label{fig:pacmen-ablation}
    }
    \vspace{-5pt}
    \caption{The performance comparison and ablation study of the Rochico algorithm.}
    \label{fig:my_label}
\end{figure*}

{\em{Pacmen}} and {\em{Block}} need to pay more attention to the division of labor of agents than {\em{Pursuit}} and {\em{Battle}} environments, so the diversity of policies and the degree of collaboration between agents should have a greater impact on the final performance.
As can be seen from Figures~\ref{fig:pacmen-totalreward} and Figures~\ref{fig:block-totalreward}, since IDQN does not take into account the cooperation between agents, the weak individual ability limits the overall task completion.
Compared with IDQN, QMIX, and NCC-Q, which encourage the cooperation between agents explicitly, achieve a certain degree of diversification of policies with better performance.
In addition, NCC-Q explicitly imposes consensus constraints on the policies within the team, thereby it can achieve better collaboration.
However, QMIX and NCC-Q both use predefined teaming strategies, they cannot dynamically adapt to the non-stationary environment.
{\sc{Rochico}} outperforms all the other three methods: it can perform adaptive teaming because of reinforced organization control; and it can achieve tight cooperation even when the teams change dynamically because of the hierarchical consensus learning.

For {\em{Pursuit}} and {\em{Battle}} environments, more attentions are paid on the flexibility of agents' collaboration policies. 
{\sc{Rochico}} can achieve the best performance in these two more complex environments, as seen from Figure~\ref{fig:pursuit-totalreward} and Figure~\ref{fig:battle-totalreward}.
The ability of the individual agent is more important in {\em{Pursuit}}, as a result, the performance of IDQN is better than QMIX and NCC-Q;
the flexible switching ability of the individual agent is more important in {\em{Battle}}, the performance of IDQN becomes poor.


\subsection{Ablation Study}
In this part, we first conduct an ablation analysis on the three important components of the {\sc{Rochico}} algorithm: hierarchical consensus constraints, structural consistency intrinsic reward in the organization control module, and the intrinsic reward in the decision module.
Here we choose the {\em{Pursuit}} environment because all comparisons have the most significant difference in this environment.
The detailed comparisons can be summarized as follows:

\noindent {\bf{1)}}. We can see from Figure ~\ref{fig:pursuit-ablation} that removing the structural consistency intrinsic reward (i.e., {\sc{Rochico-c}}) has the least impact on the algorithm performance.
This is because the edge between two nodes in the organization control module is determined by the {\sc{OR}} operation, which makes it difficult for the edge existing at the previous timestep to disappear due to randomness at the next timestep.
This indirectly realizes a certain degree of regularity for the stability of the graph structure.

\noindent {\bf{2)}}. It can be seen from the Figure ~\ref{fig:pursuit-ablation} that after removing the hierarchical consistency constraint ({\sc{Rochico-g}}, shown in Figure~\ref{fig:icg}(a)), the performance of the {\sc{Rochico}} algorithm has dropped significantly. 
Although IDQN performs better than QMIX and NCC-Q in Figure~\ref{fig:pursuit-totalreward} , this does not mean that cooperation is unnecessary in {\em{Pursuit}} environment.
QMIX and NCC-Q use a predefined teaming strategy instead of the adaptive strategy in {\sc{Rochico}}, which limits the capabilities of the agents.

\noindent{\bf{3)}}. After removing the intrinsic reward in decision module ({\sc{Rochico-i}}, replaced by sum of local rewards of all agents in the team), the performance of {\sc{Rochico-i}} has a great decline from {\sc{Rochico}}.
This shows that use pf team intention to promote diversification indirectly is difficult.
Through utilizing the intrinsic rewards based on intention difference between teams, the diversified policy emergence will become easier.

\begin{table}[ht!]
\centering
\resizebox{0.48\textwidth}{!}{%
\begin{tabular}{l|ccc||cc||c}
\hline
 &
  \multicolumn{1}{l}{\textsc{Rochico-C}} &
  \multicolumn{1}{c}{\textsc{Rochico-G}} &
  \multicolumn{1}{c||}{\textsc{Rochico-I}} &
  \multicolumn{1}{c}{\textsc{Rochico-1}} &
  \multicolumn{1}{c||}{\textsc{Rochico-3}} &
  \multicolumn{1}{c}{\textsc{Rochico(-2)}} \\ 
 \hline
{\em{Pacmen}}  & $953(\pm 56)$ & $889(\pm 78)$  & $803(\pm 45)$    & $893(\pm 52)$  & $767(\pm 74)$  & $\mathbf{988(\pm 40)}$  \\
{\em{Block}}   & $687(\pm 45)$ & $635(\pm 38)$  & $580(\pm 52)$    & $685(\pm 56)$  & $620(\pm 44)$  & $\mathbf{736(\pm 46)}$  \\
{\em{Pursuit}} & $2252(\pm 420)$ & $2012(\pm 385)$ & $1588(\pm 394)$  & $2486(\pm 403)$ & $2403(\pm 345)$ & $\mathbf{2548(\pm 380)}$ \\
{\em{Battle}}  & $221(\pm 24)$ & $199(\pm 33)$  & $169(\pm 23)$    & $194(\pm 26)$  & $160(\pm 32)$  & $\mathbf{232(\pm 27)}$ \\
\hline
\end{tabular}%
}
\caption{The average episode rewards of ablation algorithms in test environment. The mean and standard variance are calculated under $5$ random seeds.}
\label{tab:ablation}
\end{table}

The ablation analysis on the decision of the organization control module is shown in Figure~\ref{fig:pacmen-ablation}. 
The number suffix $k=\{1,2,3\}$ behind the {\sc{Rochico}} indicates that each agent needs to decide at the same time whether to form a team with the nearest $k$ agents.
$k=2$ is the default setting of {\sc{Rochico}} algorithm.
It can be seen from Figure~\ref{fig:pacmen-ablation} that {\sc{Rochico-2}}({\sc{Rochico}}) outperforms both {\sc{Rochico-1}} and {\sc{Rochico-3}}.
{\sc{Rochico-1}} has a smaller range which will weaken the connection between agents, and less effective collaboration between agents can be established; 
{\sc{Rochico-3}} has a larger range which will make the teams too fixed, and unable to flexibly adapt to the non-stationary environment.
The average episode rewards of the above ablations in the test environment are shown in Table~\ref{tab:ablation}.

\subsection{Emergence Behavior Analysis}
\begin{figure}[htbp]
  \centering
  \includegraphics[width=0.3\textwidth]{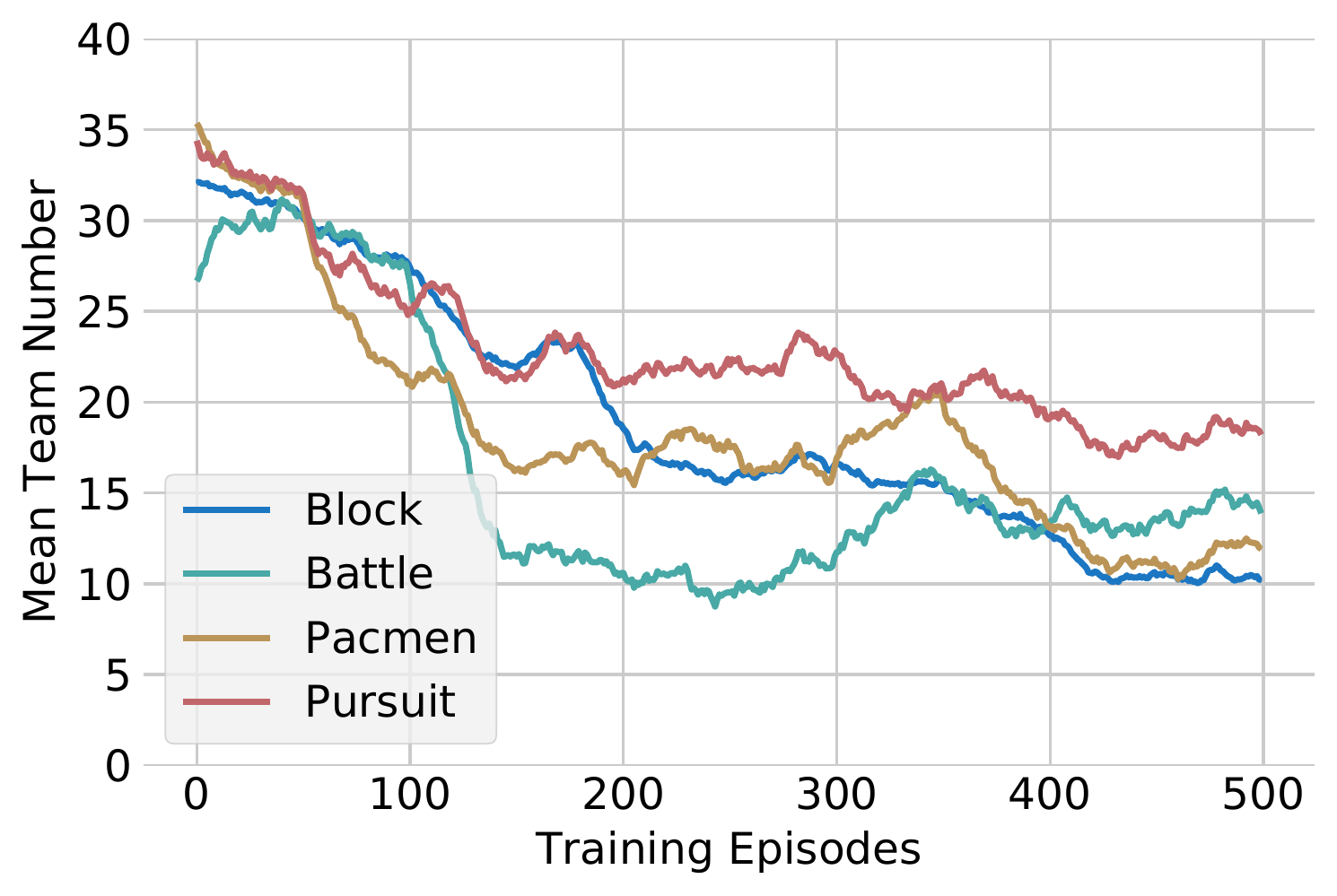}
  \caption{The averaged team number changing.}
  \label{fig:mean-teams}
\end{figure}

Figure~\ref{fig:mean-teams} shows the changing of the averaged team number in the training process.
The curves corresponding to all environments show a downward trend in the training process, which means that the organization control module does learn teaming strategies that can promote cooperation.
It can be seen from the Figure~\ref{fig:mean-teams} that the {\em{Pursuit}} environment has the largest averaged team number.
The {\em{Pursuit}} environment is more focused on the ability of the agent itself, which is also consistent with the previous analysis.

\begin{figure}[htbp]
  \centering
  \includegraphics[width=0.4\textwidth]{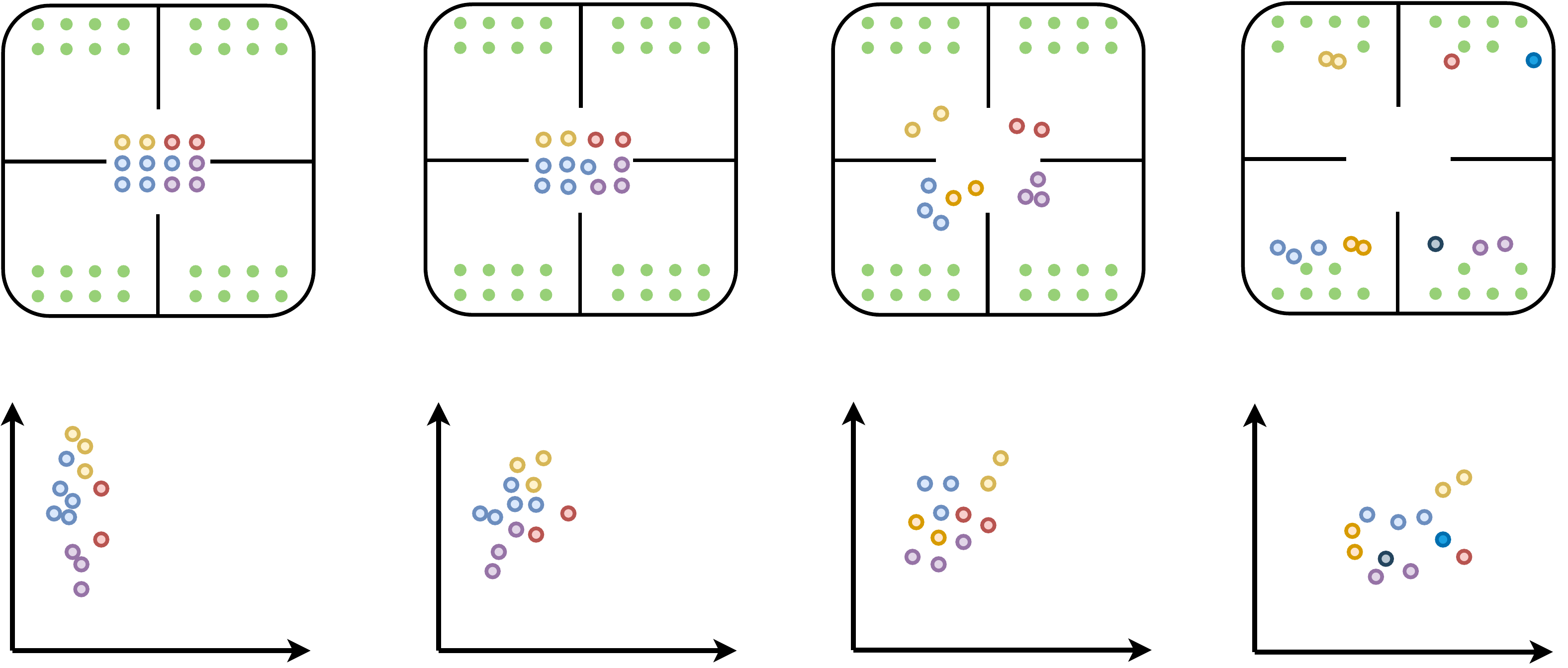}
  \caption{The relationship between team pattern and team intention in {\em{Pacmen}} environment.}
  \vspace{-10pt}
  \label{fig:pacmen-pattern}
\end{figure}

Further, we select the {\em{Pacmen}} environment to deeply analyze the teaming strategy of the organization control module and the individual intentions at different stages for task completion.
For the convenience of the presentation, we conduct the experiment in the {\em{Pacmen}} with only $12$ agents.
The different colors in Figure~\ref{fig:pacmen-pattern} represent different teams (green represents food).
The following two-dimensional scatter plot is the visualized result of the t-SNE~\citep{maaten2008visualizing} algorithm by reducing the dimensionality of the individual intentions at the corresponding timestep.
It can be seen from Figure~\ref{fig:pacmen-pattern} that the points belonging to the same team show obvious aggregation, while the points belonging to different teams are far apart. 
As the task progresses, if the difference between the sub-tasks completed by different teams becomes larger, the corresponding points will be farther away.

\section{Conclusion}
In this paper, in order to improve the efficiency of multi-agent exploration and collaboration in complex tasks, we propose a MARL framework {\sc{Rochico}} based on reinforced organization control and hierarchical consensus learning.
In the organization control module, we model the multi-agent organization control problem as a cooperative POSG and use an independent MARL algorithm to output an adaptive teaming strategy.
In the hierarchical consensus module, based on the auxiliary tasks of contrastive learning and self-supervised learning, the exploration efficiency and the collaboration efficiency of multi-agents are improved through hierarchical consensus learning. 
The comparison between {\sc{Rochico}} and current SOTA cooperative MARL algorithms in four large-scale cooperative multi-agent environments shows that our algorithm can complete complex tasks more efficiently through richer policies diversity and tighter agents collaboration.

\section{Acknowledge}
This work was supported in part by National Key Research and Development Program of China~(No. 2020AAA0107400), NSFC (No. 12071145), STCSM (No. 18DZ2270700 and 20511101100), the Open Research Projects of Zhejiang Lab (NO.2021KE0AB03) and a grant from Shenzhen Institute of Artificial Intelligence and Robotics for Society.

\bibliographystyle{ACM-Reference-Format}  
\balance
\bibliography{main}  

\newpage
\appendix
\onecolumn
\section*{\centering{Supplementary Material}}
\subsection*{A. Environments}

\noindent\textbf{Pacmen.} This scenario is a fully cooperative task, where $N$ agents initialized at the maze center and $M$ dots scatter randomly at four corner rooms.
Agents get the reward by eating dots.
Each agent has a local observation that contains a circle view with a radius $7$ centered at the agent itself.
The moving or attacking range of the agent is the only $1$ neighbor grid.
The reward is $-0.01$ for moving, $+0.5$ for attacking the dot, $-0.1$ for attacking a blank grid, and $+5$ for eat a dot.
Since the dots are distributed in different corners, the agent needs to be grouped automatically and travel to different corners to eat more dots.

\noindent\textbf{Block.} This scenario is a fully cooperative task, where $N$ blockers and $L$ blockees who have superior speed than the blocker.
There also are $M$ foods initialized on one side of the map.
Blockers and blockees are only rewarded by eating foods.
The moving or attacking range of the blockers is the $5$ neighbor grids and the blockees have larger local observed range.
The blockers have very large hit points and therefore are considered indestructible.
Blockees could be killed by blockers.
The reward is $0$ for moving, $-0.2$ for attacking, $-1$ for being killed, and $+5$ for eat one food.
Since the blockee runs faster than the blocker, the blocker needs to learn to use diverse strategies to block blockees and eat food at the same time.

\noindent\textbf{Pursuit.} This scenario is a fully cooperative task, where $N$ predators and $L$ preys who have superior speed than the predators. 
The moving or attacking range of the agent is the $4$ neighbor grids and the predator has a larger local observed range.
The reward is $0$ for moving, $-0.2$ for attacking the prey, $1$ for killing, $-1$ for being killed, and $-0.2$ for attacking a blank grid.
Since the prey runs faster than the predator, the predator needs to learn to round up through the division of labor and cooperation.

\begin{figure}[htbp]
  \centering
  \includegraphics[width=0.3\textwidth]{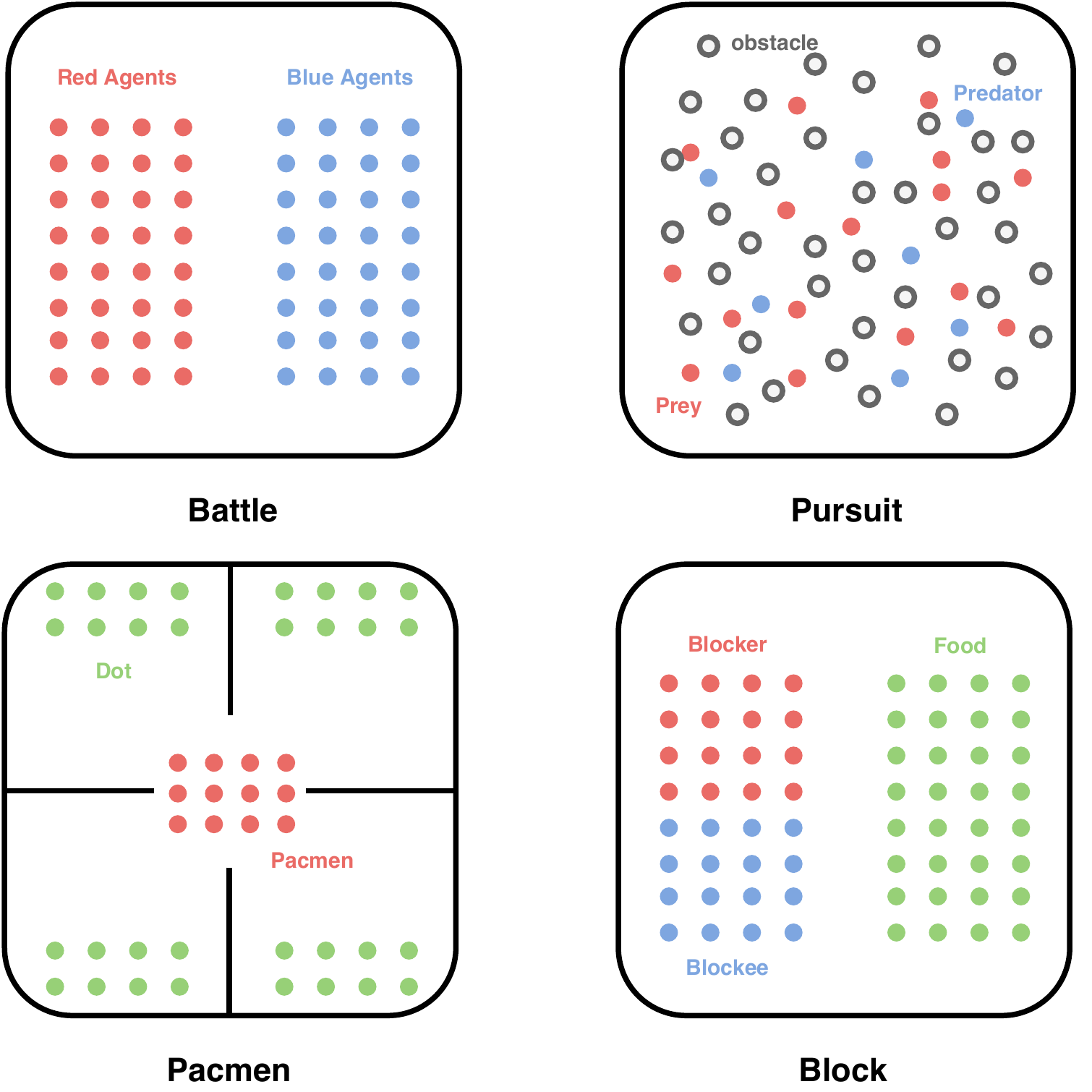}
  \vspace{-5pt}
  \caption{The simulated environments.}
  \vspace{-10pt}
  \label{fig:envs}
\end{figure}

\noindent\textbf{Battle.} This scenario is a fully cooperative task, where $N$ agents learn to fight against $L$ enemies who have superior abilities than the agents. 
The moving or attacking range of the agent is the $4$ neighbor grids, however, the enemy can move to one of $12$ nearest grids or attack one of $8$ neighbor grids. 
Each agent/enemy has $10$ hit points. 
The reward is $-0.005$ for moving, $+5$ for attacking the enemy, $-0.1$ for being killed, and $-0.1$ for attacking a blank grid. 
As the hit point of the enemy is $10$, agents have to continuously cooperate to kill the enemy. 
Therefore, the task is much more challenging than Pursuit in terms of learning to cooperate.
The schematic diagrams of all environments are shown in Figure~\ref{fig:envs}.

\subsection*{B. Hyperparameters}

The detailed hyperparameter settings of all algorithms are shown in Table~\ref{tab:hyper-rochico}, Table~\ref{tab:hyper-idqn}, Table~\ref{tab:hyper-qmix} and Table~\ref{tab:hyper-nccq}.

\begin{table}[]
\centering
\resizebox{0.48\textwidth}{!}{%
\begin{tabular}{|l|l|}
\hline
\multicolumn{1}{|c|}{\textbf{Hyperparameter}}    & \multicolumn{1}{c|}{\textbf{Value}}          \\ \hline
Batch Size                                       & $512$                        \\ \hline
Experience Replay Buffer Size                    & $1,024,000$                  \\ \hline
Target Network Update Frequency                  & $1$ times per $1000$ samples \\ \hline
Train Frequency                                  & $4$ times per batch          \\ \hline
Train Episodes                                   & $500$                        \\ \hline
Max Episode Length                               & $250$                        \\ \hline
$epsilon$ Piecewise Decay                        & $[0, 200, 400]$ episodes, {[}1, 0.2, 0.05{]} \\ \hline
Layer Number of CNN                              & $2$                          \\ \hline
Kernel Size of Each Layer                        & $3 \times 3$                 \\ \hline
Kernel Number of Each Layer                      & $32$                         \\ \hline
Hidden Size of MLP                               & $[256, 512]$                 \\ \hline
Activation Function                              & ReLU                         \\ \hline
Double Q-Learning                                & True                         \\ \hline
Dueling Q-Learning                               & True                         \\ \hline
Learning Rate                                    & $0.0001$                     \\ \hline
$\gamma$                                         & $0.99$                       \\ \hline
Hidden Sizes of  MLP in Team Intention Generator & $[32, 32, 32]$                               \\ \hline
Kernel Size of CNN in Team Intention Generator   & $3 \times 3$                 \\ \hline
Kernel Number of CNN in Team Intention Generator & $32$                         \\ \hline
Hypernet Kernel Size of CNN                      & $3 \times 3$                 \\ \hline
Hypernet Layer Number of MLP                     & $2$                          \\ \hline
Hypernet Hidden Size of MLP                      & $64$                         \\ \hline
Layer Number of GCN                              & $1$                          \\ \hline
Hidden Size of GCN                               & $32$                         \\ \hline
Hidden Sizes of VAE Encoder                      & $[32, 32]$                   \\ \hline
Hidden Sizes of VAE Decoder                      & $[32, 32]$                   \\ \hline
Dimension of Hierachical Intentions              & $32$                         \\ \hline
\end{tabular}%
}
\caption{The hyperparameters setting of $ROCHICO$ algorithm.}
\label{tab:hyper-rochico}
\end{table}

\begin{table}[]
\centering
\resizebox{0.48\textwidth}{!}{%
\begin{tabular}{|l|l|}
\hline
\multicolumn{1}{|c|}{\textbf{Hyperparameter}} & \multicolumn{1}{c|}{\textbf{Value}}          \\ \hline
Batch Size                    & $512$               \\ \hline
Experience Replay Buffer Size & $4,096,000$         \\ \hline
Target Network Update Frequency               & $1$ times per $1000$ samples                 \\ \hline
Train Frequency               & $4$ times per batch \\ \hline
Train Episodes                & $500$               \\ \hline
Max Episode Length            & $250$               \\ \hline
$epsilon$ Piecewise Decay                     & $[0, 200, 400]$ episodes, {[}1, 0.2, 0.05{]} \\ \hline
Layer Number of CNN           & $2$                 \\ \hline
Kernel Size of Each Layer     & $3 \times 3$        \\ \hline
Kernel Number of Each Layer   & $32$                \\ \hline
Hidden Size of MLP            & $[256, 512]$        \\ \hline
Activation Function           & ReLU                \\ \hline
Double Q-Learning             & True                \\ \hline
Dueling Q-Learning            & True                \\ \hline
Learning Rate                 & $0.0001$            \\ \hline
$\gamma$                      & $0.99$              \\ \hline
\end{tabular}%
}
\caption{The hyperparameters setting of $IDQN$ algorithm.}
\label{tab:hyper-idqn}
\end{table}

\begin{table}[]
\centering
\resizebox{0.48\textwidth}{!}{%
\begin{tabular}{|l|l|}
\hline
\multicolumn{1}{|c|}{\textbf{Hyperparameter}}       & \multicolumn{1}{c|}{\textbf{Value}}          \\ \hline
Batch Size                                           & $512$               \\ \hline
Experience Replay Buffer Size                        & $4,096,000$         \\ \hline
Target Network Update Frequency                     & $1$ times per $1000$ samples                 \\ \hline
Train Frequency                                      & $4$ times per batch \\ \hline
Train Episodes                                       & $500$               \\ \hline
Max Episode Length                                   & $250$               \\ \hline
$epsilon$ Piecewise Decay                           & $[0, 200, 400]$ episodes, {[}1, 0.2, 0.05{]} \\ \hline
Layer Number of CNN                                  & $2$                 \\ \hline
Kernel Size of Each Layer                            & $3 \times 3$        \\ \hline
Kernel Number of Each Layer                          & $32$                \\ \hline
Hidden Size of MLP                                   & $[256, 512]$        \\ \hline
Activation Function                                  & ReLU                \\ \hline
Double Q-Learning                                    & True                \\ \hline
Dueling Q-Learning                                   & True                \\ \hline
Learning Rate                                        & $0.0001$            \\ \hline
$\gamma$                                             & $0.99$              \\ \hline
Hypernet Kernel Size of CNN & $3 \times 3$                                 \\ \hline
Hypernet Layer Number of MLP & $2$                 \\ \hline
Hypernet Hidden Size of MLP  & $64$                \\ \hline
\end{tabular}%
}
\caption{The hyperparameters setting of $QMIX$ algorithm.}
\label{tab:hyper-qmix}
\end{table}

\begin{table}[]
\centering
\resizebox{0.48\textwidth}{!}{%
\begin{tabular}{|l|l|}
\hline
\multicolumn{1}{|c|}{\textbf{Hyperparameter}} & \multicolumn{1}{c|}{\textbf{Value}}          \\ \hline
Batch Size                    & $512$               \\ \hline
Experience Replay Buffer Size & $1,024,000$         \\ \hline
Target Network Update Frequency               & $1$ times per $1000$ samples                 \\ \hline
Train Frequency               & $4$ times per batch \\ \hline
Train Episodes                & $500$               \\ \hline
Max Episode Length            & $250$               \\ \hline
$epsilon$ Piecewise Decay                     & $[0, 200, 400]$ episodes, {[}1, 0.2, 0.05{]} \\ \hline
Layer Number of CNN           & $2$                 \\ \hline
Kernel Size of Each Layer     & $3 \times 3$        \\ \hline
Kernel Number of Each Layer   & $32$                \\ \hline
Hidden Size of MLP            & $[256, 512]$        \\ \hline
Activation Function           & ReLU                \\ \hline
Double Q-Learning             & True                \\ \hline
Dueling Q-Learning            & True                \\ \hline
Learning Rate                 & $0.0001$            \\ \hline
$\gamma$                      & $0.99$              \\ \hline
Hypernet Kernel Size of CNN   & $3 \times 3$        \\ \hline
Hypernet Layer Number of MLP  & $2$                 \\ \hline
Hypernet Hidden Size of MLP   & $64$                \\ \hline
Layer Number of GCN           & $1$                 \\ \hline
Hidden Size of GCN            & $32$                \\ \hline
Hidden Sizes of VAE Encoder   & $[32, 32]$          \\ \hline
Hidden Sizes of VAE Decoder   & $[32, 32]$          \\ \hline
Dimension of Latent Variable  & $32$                \\ \hline
\end{tabular}%
}
\caption{The hyperparameters setting of $NCC-Q$ algorithm.}
\label{tab:hyper-nccq}
\end{table}

\subsection*{C. Notations}

The notations of three module in {\sc{rochico}} are shown in Table~\ref{tab:nota-oc}, Table~\ref{tab:nota-hc} and Table~\ref{tab:nota-dec}.

\begin{table*}[]
\centering
\resizebox{0.8\textwidth}{!}{%
\begin{tabular}{|l|l|}
\hline
\multicolumn{1}{|c|}{\textbf{Symbols}}   & \multicolumn{1}{c|}{\textbf{Descriptions}}                                                   \\ \hline
$\mathcal{G}$                   & The directed graph constructed by all agents in the environment.                    \\ \hline
$\mathcal{V}$                   & The node set of graph $\mathcal{G}$ and each node represents a agent.               \\ \hline
$\mathcal{E}$                   & The edge set of graph $\mathcal{G}$ and edges are determined by agents' policies.   \\ \hline
$v$                             & A node in node set $\mathcal{V}$.                                                   \\ \hline
$n$                             & The mode of  node set $\mathcal{V}$.                                                \\ \hline
$\mathcal{M}_{org}$             & The POSG used to model organization control problem.                                \\ \hline
$\mathcal{X}_{u}$               & The agent space of  $\mathcal{M}_{org}$.                                            \\ \hline
$\mathcal{S}_{u}$               & The state space of $\mathcal{M}_{org}$.                                             \\ \hline
$\mathcal{A}_{u}^i$             & The action space of agent $i$ belongs to agent space $\mathcal{X}_{u}$.             \\ \hline
$\mathcal{O}_{u}^i$             & The obsercation space of agent $i$ belongs to agent space $\mathcal{X}_{u}$.        \\ \hline
$\mathcal{P}_{u}$               & The transition model of $\mathcal{M}_{org}$.                                        \\ \hline
$\mathcal{E}_{u}$               & The emission probability model of $\mathcal{M}_{org}$.                              \\ \hline
$\mathcal{R}_{u}^i$             & The reward function of agent $i$ belongs to agent space $\mathcal{X}_{u}$.          \\ \hline
$a_{u}^i, a_{u}^{i,\prime}$     & Two consecutive actions belongs to action space $\mathcal{A}_{u}^i$.                \\ \hline
$o_{u}^i, o_{u}^{i,\prime}$                                & The consecutive observations belongs to observation space $\mathcal{O}_{u}^i$.                   \\ \hline
$m$                             & The upper bound of nearest neighbors of any agent in agent space $\mathcal{X}_{u}$. \\ \hline
$\mathcal{N}_m(i)$              & The $m$-nearest neighbors set of agent $i$ belongs to $\mathcal{X}_{u}$.            \\ \hline
$d(i)$                          & The index of agent $i$ of action belongs to action space $\mathcal{A}_{u}^i$.       \\ \hline
$\mathcal{G}_u$                 & The corresponding undirected graph of $\mathcal{G}$.                                \\ \hline
$\mathcal{V}_u$                 & The node set of graph $\mathcal{G}_u$ and each node represents a agent.             \\ \hline
$\mathcal{E}_u$                                            & The edge set of graph $\mathcal{G}_u$ and edges are determined by agents' policies.              \\ \hline
$e(i,j)$                        & The edge between node $i$ and node $j$ in $\mathcal{G}_u$.                          \\ \hline
$r_{e}^i$                       & The external reward of agent $i$.                                                   \\ \hline
$r_{u}^i$                       & The structural consistency intrinsic reward of agent $i$.                           \\ \hline
$r_{u+}^{i}$                    & The total reward of agent $i$.                                                      \\ \hline
$\alpha_u$                      & The strength of contraint for structural consistency.                               \\ \hline
$\text{GED}(\cdot)$             & The graph edit distance                                                             \\ \hline
$\mathcal{G}_{u}\left(\mathcal{N}_{m}(i)\vee i\right)$     & The sub-graph only contains node $i$ and its $m$-nearest neighbors before take action $a_{u}^i$. \\ \hline
$\mathcal{G}_{u}^{'}\left(\mathcal{N}_{m}(i)\vee i\right)$ & The sub-graph only contains node $i$ and its $m$-nearest neighbors after take action $a_{u}^i$.  \\ \hline
$\mathcal{J}_{u}$               & The optimization goal of organization control module.                               \\ \hline
$\tau_{u}$                      & The sample trajectory.                                                              \\ \hline
$\mathcal{L}_{u}^i(\cdot)$      & The TD($0$) error of agent $i$.                                                     \\ \hline
$\theta_u, \bar{\theta_u}$      & The parameters of $Q$ function and target $Q$ function.                             \\ \hline
$D$                             & The experience replay buffer.                                                       \\ \hline
$y$                             & The TD backup.                                                                      \\ \hline
$\gamma$                        & The discount factor.                                                                \\ \hline
$\mathcal{L}_{u}^{Q}(\theta_u)$ & The overall objective function of organization control module.                      \\ \hline
\end{tabular}%
}
\caption{Notations in the organization control module.}
\label{tab:nota-oc}
\end{table*}

\begin{table*}[]
\centering
\resizebox{0.8\textwidth}{!}{%
\begin{tabular}{|l|l|}
\hline
\multicolumn{1}{|c|}{\textbf{Symbols}}        & \multicolumn{1}{c|}{\textbf{Descriptions}}                                               \\ \hline
$f_{\mu}(\cdot)$                              & The state encoder parameterized by $\mu$.                                                \\ \hline
$k, u, v$                                     & The team indicators.                                                                     \\ \hline
$n_k$                                         & The agent number of team $k$.                                                            \\ \hline
$o_{t}^{k,i}$                                 & The observation of agent $i$ in team $k$ at timestep $t$.                                \\ \hline
$\boldsymbol{o}_{t}^{k}$                      & The joint observation of all agents in team $k$ at timestep $t$.                         \\ \hline
$e_{t}^{k,i}$                                 & The observation embedding of agent $i$ in team $k$ at timestep $t$.                      \\ \hline
$f_{\nu}(\cdot)$                              & The DeepSet network parameterized by $\nu$.                                              \\ \hline
${e}_t^{k}$                                   & The team embedding of team $k$ at timestep $t$.                                          \\ \hline
$f_{\omega}(\cdot)$                           & The team intention encoder parameterized by $\omega$.                                    \\ \hline
$s_t$                                         & The global state of timestep $t$.                                                        \\ \hline
${c}_t^{k}$                                   & The team intention of team $k$ at timestep $t$.                                          \\ \hline
$(x_t^{k,i}, y_t^{k,i})$                      & The spatial position of agent $i$ in team $k$ at timestep $t$.                           \\ \hline
$[\bar{x}_t^k, \bar{y}_t^k]$                  & The average spatial position of team $k$ at timestep $t$.                                \\ \hline
$t_k$                                         & The timestep of team $k$.                                                                \\ \hline
$e_{ts}^{k}$                                  & The spatiotemporal feature representation of team $k$.                                   \\ \hline
$d(k,l)$                                      & The Euclidean distance in the spatiotemporal space between team $k$ and $l$.             \\ \hline
$\mathcal{G}_{\ell}({\mathcal{V}}_{\ell}, {\mathcal{E}}_{\ell})$ &
  \begin{tabular}[c]{@{}l@{}}The undirected graph constructed by team \\ (similar as undirected graph in organization control module).\end{tabular} \\ \hline
$e(k,l)$                                      & The edge between team $k$ and team $l$.                                                  \\ \hline
$y_t^k$                                       & The label of team $k$.                                                                   \\ \hline
$\mathcal{L}^{k}_\ell(\mu,\nu,\omega)$ &
  The contrasive learning objective function of hierachical consensus learning module. \\ \hline
$m$                                           & The margin parameter in contrasive learning objective function.                          \\ \hline
$f_{\xi}(\cdot)$                              & The team intention decoder parameterized by $\xi$.                                       \\ \hline
$\hat{o}_{t+1}^{k,i}$ &
  \begin{tabular}[c]{@{}l@{}}The reconstructed observation by team intention decoder \\ of agent $i$ in team $k$ at timestep $t$.\end{tabular} \\ \hline
$\boldsymbol{\hat{o}}_{t+1}^{k}$ &
  \begin{tabular}[c]{@{}l@{}}The joint reconstructed observation by team intention decoder \\ of all agents in team $k$ at timestep $t$.\end{tabular} \\ \hline
$\mathcal{L}^k_{\ell}(\xi)$                   & The prediction loss of team intention decoder.                                           \\ \hline
$\mathcal{L}_{\ell}^{tg}(\mu,\nu,\omega,\xi)$ & The loss function of team intention generator.                                           \\ \hline
$\lambda_{tg}$                                & The temperature parameter of loss function of team intention generator.                  \\ \hline
$g_{\phi}(\cdot)$                             & The individual encoder parameterized by $\phi$                                           \\ \hline
$h_{t}^{k,i}$                                 & The individual embedding of agent $i$ in team $k$ at timestep $t$.                       \\ \hline
$g_{\psi}(\cdot)$                             & The graph convolutional network parameterized by $\psi$.                                 \\ \hline
$\boldsymbol{h}_{t}^k$                        & The joint embedding of all agents in team $k$ at timestep $t$.                           \\ \hline
$\chi_{t}^{k,i}$                              & The individual cognition of agent $i$ in team $k$ at timestep $t$.                       \\ \hline
$g_{\varphi}(\cdot)$                          & The variational encoder parameterized by $\varphi$.                                      \\ \hline
$\zeta_{t}^{k,i}$                             & The individual intention of agent $i$ in team $k$ at timestep $t$.                       \\ \hline
$\mathcal{L}^i_{\ell}(\phi,\psi,\varphi)$     & The hierachical consensus loss of hierachical consensus module.                          \\ \hline
$\mathcal{L}^i_{\ell}(\varphi)$               & The loss function of variation autoencoder.                                              \\ \hline
$p(\zeta_{t}^{k,i})$                          & The prior distribution of individual intention of agent $i$ in team $k$ at timestep $t$. \\ \hline
$\mathcal{L}_{\ell}^{ig}(\phi,\psi,\varphi)$  & The loss function of individual intention generator.                                     \\ \hline
\end{tabular}%
}
\caption{Notations in the hierachical consensus module.}
\label{tab:nota-hc}
\end{table*}

\begin{table*}[]
\centering
\resizebox{0.8\textwidth}{!}{%
\begin{tabular}{|l|l|}
\hline
\multicolumn{1}{|c|}{\textbf{Symbols}}                        & \multicolumn{1}{c|}{\textbf{Descriptions}}                                        \\ \hline
$r_{t,\ell}^{k}$                                              & The intrinsic reward of team $k$ at timestem $t$.                                 \\ \hline
$\mathcal{L}_{\ell}^i(\theta^i_\ell)$                         & The TD($0$) error of local $Q$ function of agent $i$ in team $k$ at timestep $t$. \\ \hline
$Q_{\theta^i_\ell}(\cdot), Q_{\bar{\theta_\ell^i}}$ &
  \begin{tabular}[c]{@{}l@{}}The local $Q$ function and local target $Q$ function of agent $i$ in team $k$,\\ parameterized by $\theta^i_\ell$ and $\bar{\theta^i_\ell}$ respectively.\end{tabular} \\ \hline
$y^i$                                                         & The TD backup of agent $i$ in team $k$.                                           \\ \hline
$\mathcal{L}_{\ell}^k(\theta^k_\ell)$                         & The TD($0$) error of joint $Q$ function of  team $k$ at timestep $t$.             \\ \hline
$Q_{\theta^k_\ell}(\cdot), Q_{\bar{\theta_\ell^k}}$ &
  \begin{tabular}[c]{@{}l@{}}The joint $Q$ function and joint target $Q$ function of team $k$,\\ parameterized by $\theta^k_\ell$ and $\bar{\theta^k_\ell}$ respectively.\end{tabular} \\ \hline
$y^k$                                                         & The TD backup of team $k$.                                                        \\ \hline
$\mathcal{L}_{\ell}^{Q}(\{\theta_\ell^i\},\{\theta_\ell^k\})$ & The loss function of decision module.                                             \\ \hline
$\lambda_{QMIX}$                                       & The temperature paramter of loss function of decision module.                     \\ \hline
\end{tabular}%
}
\caption{Notations in the decision module.}
\label{tab:nota-dec}
\end{table*}

\end{document}